\documentclass{article}
\pdfoutput=1

\usepackage{microtype}
\usepackage{graphicx}
 \graphicspath{{figures/}} 
\usepackage{subfigure}
\usepackage{booktabs} 

\usepackage{hyperref}

\usepackage[accepted]{icml2024}

\usepackage{amsmath}
\usepackage{amssymb}
\usepackage{mathtools}
\usepackage{amsthm}
\usepackage{bbm}
\usepackage{nicefrac}

\usepackage{xcolor} 
 \definecolor{mygreen}{rgb}{0,0.6,0}
 \definecolor{mygray}{rgb}{0.5,0.5,0.5}
 \definecolor{mymauve}{rgb}{0.58,0,0.82}
 \definecolor{backcolour}{rgb}{0.95,0.95,0.92}

\usepackage{listings}

\usepackage{colortbl}

\usepackage[capitalize,noabbrev]{cleveref}

\theoremstyle{plain}
\newtheorem{theorem}{Theorem}[section]

\newtheorem{lemma}[theorem]{Lemma}

\theoremstyle{definition}

\newtheorem{assumption}[theorem]{Assumption}
\theoremstyle{remark}

\usepackage[textsize=tiny]{todonotes}

\icmltitlerunning{Cross-Domain Policy Adaptation by Capturing Representation Mismatch}

\begin{document}

\twocolumn[
\icmltitle{Cross-Domain Policy Adaptation by Capturing Representation Mismatch}




\begin{icmlauthorlist}
\icmlauthor{Jiafei Lyu}{tsinghua}
\icmlauthor{Chenjia Bai}{shanghai}
\icmlauthor{Jingwen Yang}{tencent}
\icmlauthor{Zongqing Lu}{peking,zhiyuan}
\icmlauthor{Xiu Li}{tsinghua}
\end{icmlauthorlist}

\icmlaffiliation{tsinghua}{Tsinghua Shenzhen International Graduate School, Tsinghua University}
\icmlaffiliation{shanghai}{Shanghai Artificial Intelligence Laboratory}
\icmlaffiliation{tencent}{Tencent IEG}
\icmlaffiliation{zhiyuan}{Beijing Academy of Artificial Intelligence}
\icmlaffiliation{peking}{School of Computer Science, Peking University}

\icmlcorrespondingauthor{Xiu Li}{li.xiu@sz.tsinghua.edu.cn}

\icmlkeywords{Machine Learning, ICML}

\vskip 0.3in
]



\printAffiliationsAndNotice{}  

\begin{abstract}
It is vital to learn effective policies that can be transferred to different domains with dynamics discrepancies in reinforcement learning (RL). In this paper, we consider dynamics adaptation settings where there exists dynamics mismatch between the source domain and the target domain, and one can get access to sufficient source domain data, while can only have limited interactions with the target domain. Existing methods address this problem by learning domain classifiers, performing data filtering from a value discrepancy perspective, etc. Instead, we tackle this challenge from a \emph{decoupled representation learning} perspective. We perform representation learning only in the target domain and measure the representation deviations on the transitions from the source domain, which we show can be a signal of dynamics mismatch. We also show that representation deviation upper bounds performance difference of a given policy in the source domain and target domain, which motivates us to adopt representation deviation as a reward penalty. The produced representations are not involved in either policy or value function, but only serve as a reward penalizer. We conduct extensive experiments on environments with kinematic and morphology mismatch, and the results show that our method exhibits strong performance on many tasks. Our code is publicly available at \href{https://github.com/dmksjfl/PAR}{https://github.com/dmksjfl/PAR}.
\end{abstract}

\section{Introduction}
\label{sec:introduction}
Alice is interested in learning cooking. She bought a new set of cookware recently that is different from the one she used before. She expertly uses new cookware soon. As this example conveys, human beings are able to quickly transfer the learned policies to similar tasks. Such capability is also expected in reinforcement learning (RL) agents. Unfortunately, RL algorithms are known to require a vast number of interactions to learn meaningful policies \cite{Silver2016MasteringTG,Lyu2023OffPolicyRA}. A bare fact is that sometimes only limited interactions with the environment (\emph{target domain}) are feasible because it may be expensive and time-consuming for a large number of interactions in scenarios like robotics \cite{Cutler2015EfficientRL,Kober2013ReinforcementLI}, autonomous driving \cite{Kiran2020DeepRL,Osinski2019SimulationBasedRL}, etc. Nevertheless, we may simultaneously have access to another structurally similar \emph{source domain} where the experience is cheaper to gather, \emph{e.g.}, a simulator. Since the source domain can be biased, a dynamics mismatch between the two domains may persist. It then necessitates developing algorithms that have good performance in the target domain, given the source domain with some dynamics discrepancies. 

Note that there are numerous studies concerning policy adaptation, such as system identification \cite{Yu2017PreparingFT,Clavera2018LearningTA} and domain randomization \cite{Slaoui2019RobustDR,Tobin2017DomainRF,Peng2017SimtoRealTO}. These methods often rely on demonstrations from the target domain \cite{Kim2019DomainAI}, the distributions from which the simulator parameters are sampled, a manipulable simulator \cite{Chebotar2018ClosingTS}, etc. We lift these requirements and consider learning policies with sufficient source domain data (either online or offline) and limited online interactions with the target domain. This setting is also referred to as \emph{off-dynamics RL} \cite{eysenbach2021offdynamics} or \emph{online dynamics adaptation} \cite{Xu2023CrossDomainPA}. Existing methods tackle this problem by learning domain classifiers \cite{eysenbach2021offdynamics}, filtering source domain data that share similar value estimates with target domain data \cite{Xu2023CrossDomainPA}, etc.

In this paper, we study the cross-domain policy adaptation problem where only transition dynamics between the source domain and the target domain differ. The state space, action space, as well as the reward function, are kept unchanged. Unlike prior works, we address this issue from a representation learning perspective. Our motivation is that the dynamics mismatch between the source domain and the target domain can be captured by representation deviations of transitions from the two domains, which is grounded by our theoretical analysis. We further show concrete performance bounds given either online or offline source domain, where we observe that representation deviation upper bounds the performance difference of any given policy between the source domain and the target domain. Motivated by the theoretical findings, we deem that representation mismatch between two domains can be used as a reward penalizer to fulfill dynamics-aware policy adaptation.

\begin{figure}
    \centering
    \includegraphics[width=0.95\linewidth]{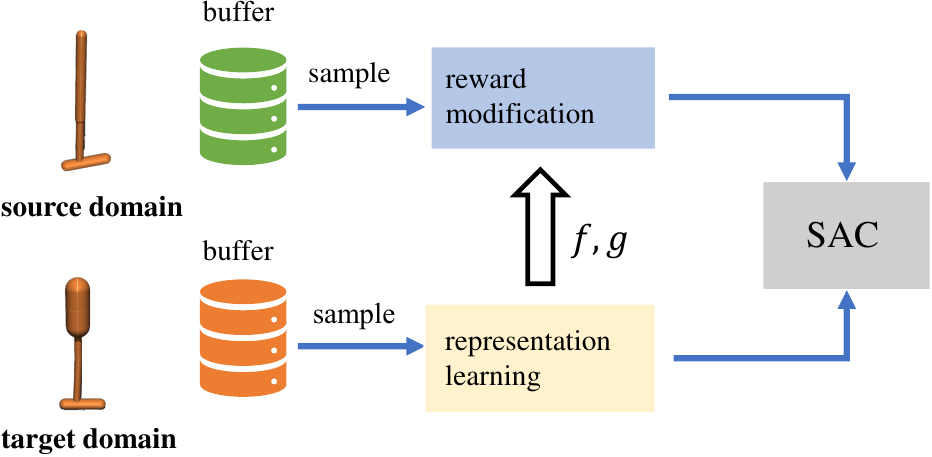}
    \caption{\textbf{Illustration of PAR}. We train encoders $f,g$ merely with target domain data and utilize them to modify rewards from the source domain with measured representation deviations. Afterward, the downstream SAC algorithm can learn from transitions from both domains.}
    \label{fig:pipeline}
\end{figure}

For a practical usage, we propose \textbf{P}olicy \textbf{A}daptation by \textbf{R}epresentation mismatch, dubbed \textbf{PAR} algorithm. Our approach trains a state encoder and a state-action encoder \emph{only in the target domain} to capture its latent dynamics structure, and then leverage the learned encoders to produce representations upon \emph{transitions from the source domain}. We evaluate deviations between representations of the state-action pair and the next state, and use the resulting representation deviations to penalize source domain rewards, as depicted in Figure \ref{fig:pipeline}. Intuitively, the penalty is large if the transition deviates far from the target domain, and vice versa. In this way, the agent can benefit more from dynamics-consistent transitions and de-emphasize others. It is worth noting that the representation learning is decoupled from policy or value function training since the representations are not involved in them. Empirical results in environments with kinematic and morphology shifts show that our method notably beats previous strong baselines on many tasks in both online and offline source domain settings.

\section{Related Work}
\label{sec:relatedwork}

\textbf{Domain Adaptation in RL}. Generalizing or transferring policies across varied domains remains a critical issue in RL, where domains may differ in terms of agent embodiment \cite{Liu2022REvolveRCE,zhang2021learning}, transition dynamics \cite{eysenbach2021offdynamics,Viano2020RobustIR}, observation space \cite{Gamrian2018TransferLF,Bousmalis2018UsingSA,Ge2022PolicyAF,Zhang2021PolicyTA,hansen2021selfsupervised}, etc. We focus on policy adaptation under dynamics discrepancies between the two domains. Prior works mainly address this issue via system identification \cite{Clavera2018LearningTA,zhou2018environment,Du2021AutoTunedST,Xie2022RobustPL}, domain randomization \cite{Slaoui2019RobustDR,Mehta2019ActiveDR,Vuong2019HowTP,Jiang2023VarianceRD}, meta-RL \cite{nagabandi2018learning,Raileanu2020FastAT,Arndt2019MetaRL,Wu2022ZeroShotPT}, or by leveraging expert demonstrations from the target domain \cite{Kim2019DomainAI,Hejna2020HierarchicallyDI,fickinger2022crossdomain,raychaudhuri2021cross}. Though effective, these methods depend on a model of the environment, expert trajectories gathered in the target domain, or a proper choice of randomized parameters. In contrast, we dismiss these requirements and study dynamics adaptation problem \cite{Xu2023CrossDomainPA} where only a small amount of online interactions with the target domain is allowed, and a source domain with sufficient data can be accessed. Under this setting, many approaches have been developed, such as directly optimizing the parameters of the simulator to calibrate the dynamics of the source domain \cite{Farchy2013HumanoidRL,Zhu2017FastMI,Collins2020TraversingTR,Chebotar2018ClosingTS,Ramos2019BayesSimAD}. However, it requires a manipulable simulator. There are attempts to use some expressive models to learn the dynamics change \cite{Golemo2018SimtoRealTW,Hwangbo2019LearningAA,xiong2023universal}, and action transformation methods that learn dynamics models of the two domains and utilize them to modify transitions from the source domain \cite{Hanna2021GroundedAT,desai2020imitation}, while it is difficult to learn accurate dynamics models \cite{Malik2019CalibratedMD,lyu2022double}. Another line of research trains domain classifiers and tries to close the dynamics gap by either reward modification \cite{eysenbach2021offdynamics,liu2022dara}, or importance weighting \cite{niu2022when}. A recent work \cite{Xu2023CrossDomainPA} bridges the dynamics gap by selectively sharing transitions from the source domain that have similar value estimates as those in the target domain. Unlike these methods, we capture dynamics discrepancy by measuring representation mismatch. It is worth noting that in this work we only consider policy adaptation across domains that have the same state space and action space. Our method can also generalize to the setting where target domain has a different state space or action space by incorporating extra components or modules like prior works \cite{barekatain2019multipolar,you2022cross,gui2023cross}. 

\textbf{Representation Learning in RL}. Representation learning is an important research topic in computer vision \cite{Bengio2012RepresentationLA,Kolesnikov2019BigT,He2015DeepRL}. In the context of RL, representation learning is actively explored in image-based tasks \cite{Kostrikov2020ImageAI,yarats2022mastering,liu2021returnbased,Cetin2022StabilizingOD}, aiming at extracting useful features from information-redundant images by contrastive learning \cite{Srinivas2020CURLCU,Eysenbach2022ContrastiveLA,stooke2021decoupling,Zhu2020MaskedCR}, MDP homomorphisms \cite{Pol2020MDPHN,rezaei-shoshtari2022continuous}, bisimulation \cite{ferns2011bisimulation,zhang2021learninginvariant}, self-predictive learning \cite{schwarzer2021dataefficient,Tang2022UnderstandingSL,Kim2022SelfPredictiveDF}, etc. Representation learning can also be found in model-based RL methods that rely on latent dynamics \cite{Karl2016DeepVB,Rafailov2020OfflineRL,hafner2019learning,Hansen2022TemporalDL}. In state-based tasks, it also spans in successor representation \cite{Barreto2016SuccessorFF,Fujimoto2021ADR,machado2023temporal}, learning state-action representations \cite{ota2020can,Fujimoto2023ForSS}, action representations \cite{Whitney2020Dynamics-Aware,Chandak2019LearningAR} for improving sample efficiency, etc. We capture latent dynamics information by learning state-action representations, but we differ from previous approaches in that we use them for detecting dynamics mismatch.

\section{Preliminaries}
\label{sec:preliminary}

We formulate reinforcement learning (RL) problems as a Markov Decision Process (MDP), which can be specified by the 5-tuple $\mathcal{M} = ( \mathcal{S},\mathcal{A}, P, r, \gamma )$, where $\mathcal{S}$ is the state space, $\mathcal{A}$ is the action space, $P$ denotes the transition dynamics, $r: \mathcal{S}\times\mathcal{A}\rightarrow \mathbb{R}$ is the scalar reward signal, and $\gamma\in[0,1)$ is the discount factor. The objective of RL is to find a policy $\pi:\mathcal{S}\rightarrow \Delta(\mathcal{A})$ that maximize the discounted cumulative return $\sum_{t=0}^\infty \gamma^t r(s_t,a_t)$. We consider access to a source domain $\mathcal{M}_{\rm src}=( \mathcal{S},\mathcal{A}, P_{\rm src}, r, \gamma )$ and a target domain $\mathcal{M}_{\rm tar}=( \mathcal{S},\mathcal{A}, P_{\rm tar}, r, \gamma )$ that share the state space and action space, and only differ in their transition dynamics. We assume the rewards are bounded, \emph{i.e.}, $|r(s,a)|\le r_{\rm max},\forall\, s,a$. 

In the rest of the paper, we specify the transition dynamics in a domain $\mathcal{M}$ as $P_{\mathcal{M}}$ (\emph{e.g.}, $P_{\mathcal{M}_{\rm src}}$ is the transition dynamics in the source domain). We denote $\rho_\mathcal{M}^\pi(s,a):=(1-\gamma)\sum_{t=0}^\infty \gamma^t P_{\mathcal{M},t}^\pi(s) \pi(a|s)$ as the normalized probability that a policy $\pi$ encounters the state action pair $(s,a)$, and $P_{\mathcal{M},t}^\pi(s)$ is the probability that the policy $\pi$ encounters the state $s$ at timestep $t$ in the domain $\mathcal{M}$. The expected return of a policy $\pi$ in MDP $\mathcal{M}$ can then be simplified as $J_{\mathcal{M}}(\pi) = \mathbb{E}_{s,a\sim\rho_{\mathcal{M}}^\pi}[r(s,a)]$.

\textbf{Notations}: $I(X;Y)$ denotes the mutual information between two random variables $X,Y$. $\mathbb{H}(X)$ is the entropy of the random variable $X$. $\Delta$ is the probability simplex.

\section{Dynamics Adaptation by Representation Mismatch}
\label{sec:method}

In this section, we start by theoretically unpacking the equivalence between the representation mismatch and the dynamics mismatch. We further show the performance bounds of a policy between the target domain and either online or offline source domain, where representation mismatch serves as the lower bound of the performance difference. Empowered by theoretical results, we leverage the representation mismatch to penalize source domain data and propose our practical algorithm for dynamics-aware policy adaptation. 


\subsection{Theoretical Analysis}

Before moving to our theoretical results, we need to impose the following assumption, which can be generally satisfied in practice (\emph{e.g.}, deep RL). We defer the detailed discussion on the rationality of this assumption to Section \ref{sec:rationalityassumption}.

\begin{assumption}[One-to-one Representation Mapping]
\label{ass:1}
    For any state-action pair $(s,a)$ and its latent representation $z$, they construct a one-to-one mapping from the original state-action joint space $\mathcal{S}\times\mathcal{A}$ to the latent space $\mathcal{Z}$.
\end{assumption}

Our first result in Theorem \ref{theo:mutual} establishes a connection between mutual information and the representation deviation of transitions from different domains. Due to space limits, all proofs are deferred to Appendix \ref{sec:missingproofs}. 

\begin{theorem}
\label{theo:mutual}
    For any $(s,a)$, denote its representation as $z$, and suppose $s_{\rm src}^\prime\sim P_{\mathcal{M}_{\rm src}}(\,\cdot\,|s,a)$, $s_{\rm tar}^\prime\sim P_{\mathcal{M}_{\rm tar}}(\,\cdot\,|s,a)$. Denote $h(z;s_{\rm src}^\prime,s_{\rm tar}^\prime)=I(z;s_{\rm tar}^\prime) - I(z;s_{\rm src}^\prime)$, then we have measuring $h(z;s_{\rm src}^\prime,s_{\rm tar}^\prime)$ is equivalent to measuring the representation deviation $D_{\rm KL}(P(z|s_{\rm tar}^\prime) \| P(z|s_{\rm src}^\prime))$.
\end{theorem}
\textbf{\emph{Remark}.} The defined function $h(z;s_{\rm src}^\prime,s_{\rm tar}^\prime)$ measures the difference between the embedded target domain information and source domain information in $z$. This theorem illustrates that such a difference is equivalent to the KL-divergence between the distributions of $z$ given source domain state and target domain state, respectively. Intuitively, $h(z;s_{\rm src}^\prime,s_{\rm tar}^\prime)$ approaches 0 if the distribution of $s_{\rm src}^\prime$ is close to that of $s_{\rm tar}^\prime$. If we enforce $z$ to contain only target domain knowledge, $h(z;s_{\rm src}^\prime,s_{\rm tar}^\prime)$ can be large if the dynamics mismatch between data from the two domains is large, incurring a large $D_{\rm KL}(P(z|s_{\rm tar}^\prime) \| P(z|s_{\rm src}^\prime))$. Naturally, one may think of using this representation deviation term as evidence of dynamics mismatch. 

Below, we show that the representation deviation can \emph{strictly} reflect the dynamics discrepancy between two domains.
\begin{theorem}
\label{theo:equal}
    Measuring the representation deviation between the source domain and the target domain is equivalent to measuring the dynamics mismatch between two domains. Formally, we can derive that $D_{\rm KL}\left( P(z|s_{\rm tar}^\prime) \| P(z|s_{\rm src}^\prime) \right) = D_{\rm KL}\left( P(s_{\rm tar}^\prime|z) \| P(s_{\rm src}^\prime|z) \right) + \mathbb{H}(s_{\rm tar}^\prime) - \mathbb{H}(s_{\rm src}^\prime).$
\end{theorem}
The above theorem conveys the rationality of detecting dynamics shifts with the aid of the representation mismatch. This is appealing as representations can contain rich information and capture hidden features, and learning in the latent space is effective \cite{Hansen2022TemporalDL}. To see how representation mismatch affects the performance of the agent, we derive a novel performance bound of a policy given online target domain and \emph{online source domain} in Theorem \ref{theo:online}.

\begin{theorem}[Online performance bound]
    \label{theo:online}
    Denote $\mathcal{M}_{\rm src}$, $\mathcal{M}_{\rm tar}$ as the source domain and the target domain, respectively, then the return difference of any policy $\pi$ between $\mathcal{M}_{\rm src}$ and $\mathcal{M}_{\rm tar}$ is bounded:
    \begin{align*}
        &J_{\mathcal{M}_{\rm tar}}(\pi) - J_{\mathcal{M}_{\rm src}}(\pi) \ge \\
        &- \dfrac{\sqrt{2}\gamma r_{\rm max}}{(1-\gamma)^2}\underbrace{\mathbb{E}_{\rho_{\mathcal{M}_{\rm src}}^\pi}\left[\sqrt{D_{\rm KL}(P(z|s_{\rm src}^\prime) \| P(z|s_{\rm tar}^\prime))}\right]}_{(a): \rm \,representation\, mismatch} \\
        &- \dfrac{\sqrt{2}\gamma r_{\rm max}}{(1-\gamma)^2} \underbrace{\mathbb{E}_{\rho_{\mathcal{M}_{\rm src}}^\pi}\left[\sqrt{|\mathbb{H}(s_{\rm src}^\prime) - \mathbb{H}(s_{\rm tar}^\prime)|}\right]}_{(b): \rm \,state\, distribution\, deviation}. 
    \end{align*}
\end{theorem}
\textbf{\emph{Remark}.} The above bound indicates that the performance difference of a policy $\pi$ in different domains is decided by the representation mismatch term (a), and the state distribution deviation term (b). Since both two domains are fixed, the entropy of their state distributions are constants, and term (b) is also a constant accordingly. Term (b) characterizes the inherent performance difference of a policy in two domains and vanishes if the two domains are identical.

Moreover, if the source domain is \emph{offline} (\emph{i.e.}, one can only have access to a static offline source domain dataset), we can derive a similar bound as shown below.

\begin{theorem}[Offline performance bound]
\label{theo:offline}
    Denote the empirical policy distribution in the offline dataset $D$ from source domain $\mathcal{M}_{\rm src}$ as $\pi_D:=\frac{\sum_{D}\mathbbm{1}(s,a)}{\sum_{D}\mathbbm{1}(s)}$, then the return difference of any policy $\pi$ between the source domain $\mathcal{M}_{\rm src}$ and the target domain $\mathcal{M}_{\rm tar}$ is bounded:
    \begin{align*}
        &J_{\mathcal{M}_{\rm tar}}(\pi) - J_{\mathcal{M}_{\rm src}}(\pi) \ge \\
        & - \dfrac{4r_{\rm max}}{(1-\gamma)^2}\underbrace{\mathbb{E}_{\rho_{\mathcal{M}_{\rm src}}^{\pi_D},P_{\mathcal{M}_{\rm src}}}[D_{\rm TV}(\pi_D || \pi )]}_{(a):\, \rm policy\, deviation} \\
        &- \dfrac{\sqrt{2}\gamma r_{\rm max}}{(1-\gamma)^2}\underbrace{\mathbb{E}_{\rho_{\mathcal{M}_{\rm src}}^{\pi_D}} \left[\sqrt{D_{\rm KL}(P(z|s_{\rm src}^\prime) \| P(z|s_{\rm tar}^\prime))}\right]}_{(b):\, \rm representation\, mismatch} \\
        &- \dfrac{\sqrt{2}\gamma r_{\rm max}}{(1-\gamma)^2} \underbrace{\mathbb{E}_{\rho_{\mathcal{M}_{\rm src}}^{\pi_D}} \left[\sqrt{|\mathbb{H}(s_{\rm src}^\prime) - \mathbb{H}(s_{\rm tar}^\prime)|}\right]}_{(c):\, \rm state\, distribution\, deviation}. 
    \end{align*}
\end{theorem}
\textbf{\emph{Remark}.} This theorem also explicates the importance of the representation mismatch term (b) as a lower bound, similar to Theorem \ref{theo:online}, but it additionally highlights the role of the policy deviation term (a). It is evident that controlling the policy deviation term counts with an offline source domain.


Theorem \ref{theo:online} and \ref{theo:offline} motivate us to use the representation mismatch term as a reward penalty to encourage dynamics-consistent transitions, because it turns out that the core factor that affects the bound either with an online or offline source domain is the representation mismatch term.

\subsection{Practical Algorithm}

To acquire representations of the transitions, we train a state encoder $f_{\psi}(s)$ parameterized by $\psi$ to produce $z_1$, the representation of the state $s$, along with a state-action encoder $g_{\xi}(z, a)$ parameterized by $\xi$ that receives the state representation $z_1$ and action as inputs and outputs state-action representation $z_2$. By letting $z_2$ be close to the representation of the next state, we realize the latent dynamics consistency \cite{Hansen2022TemporalDL,Ye2021MasteringAG}. The objective function for learning these encoders gives:
\begin{equation}
\label{eq:representationobjective}
    \mathcal{L}(\psi, \xi) = \mathbb{E}_{(s,a,s^\prime)\sim D}\left[(g_{\xi}(f_\psi(s),a) - \texttt{SG}(f_\psi(s^\prime)) )^2 \right],
\end{equation}
where $D$ is the replay buffer, and $\texttt{SG}$ denotes stop gradient operator. Similar objectives are adopted in prior works \cite{ota2020can,Fujimoto2023ForSS}. A central difference is, that we only use the representations for measuring representation mismatch, instead of involving them in policy or value function training. One can also utilize a distinct objective, as long as it can embed the latent dynamics information (see Section \ref{sec:discussion}). It is worth noting that both $f$ and $g$ are deterministic to fulfill Assumption \ref{ass:1}.

Given insights from Theorem \ref{theo:mutual}, we deem that the representations ought to embed more information of the target domain, and de-emphasize the source domain knowledge, such that the representation deviations can be a better proxy of dynamics shifts. This prompts us to train the state encoder and the state-action encoder \emph{only in the target domain}, and evaluate the representation deviations upon \emph{samples from the source domain}. We then penalize the source domain rewards with the calculated deviations, \emph{i.e.}, for any transition $(s_{\rm src},a_{\rm src},r_{\rm src}, s_{\rm src}^\prime)$ from the source domain, we modify its reward to
\begin{equation}
\label{eq:rewardmodification}
    \hat{r}_{\rm src} = r_{\rm src} - \beta \times \left[g_\xi(f_\psi(s_{\rm src}),a_{\rm src}) - f_\psi(s_{\rm src}^\prime) \right]^2,
\end{equation}
where $\beta\in\mathbb{R}$ is a hyperparameter. This penalty generally captures the representation mismatch between the source domain and the target domain. $g_\xi(f_\psi(s_{\rm src}),a_{\rm src})$ represents the state-action representation in the target domain since these domains share state space and action space, and $f,g$ only encode target domain information. It approaches the representation of $s_{\rm tar}^\prime$ that is incurred by $(s_{\rm src},a_{\rm src})$ in the target domain. $f_\psi(s_{\rm src}^\prime)$, instead, denotes the representation of $s_{\rm src}^\prime$ from the source domain. A larger penalty will be allocated if the source domain data deviates too much from the dynamics of the target domain, and vice versa. Consequently, the agent can focus more on dynamics-consistent transitions and achieve better performance. Hence, such a penalty matches our theoretical results.

Formally, we introduce our novel method for cross-domain policy adaptation, \textbf{P}olicy \textbf{A}daptation by \textbf{R}epresentation Mismatch, tagged \textbf{PAR} algorithm. We use SAC \cite{haarnoja2018softactorcritic} as the base algorithm, and aim at training value functions (\emph{a.k.a.}, critics) $Q_{\theta_1}(s,a), Q_{\theta_2}(s,a)$ parameterized by $\theta_1,\theta_2$, and policy (\emph{a.k.a.}, actor) $\pi_\phi(s)$ parameterized by $\phi$. Denote $D_{\rm src}, D_{\rm tar}$ as the replay buffers of the source domain and the target domain, and let the rewards in $D_{\rm src}$ be corrected as $\hat{r}_{\rm src}$, then the objective function for training the value functions gives:
\begin{equation}
\label{eq:criticloss}
    \mathcal{L}_{\rm critic} = \mathbb{E}_{(s,a,r,s^\prime)\sim D_{\rm src}\cup D_{\rm tar}}\left[(Q_{\theta_i}(s,a) - y)^2\right],
\end{equation}
where $i\in\{1,2\}$ and $y$ is the target value, which gives
\begin{equation}
    y = r + \gamma\left[\min_{i=1,2}Q_{\theta_i^\prime}(s^\prime,a^\prime) - \alpha \log\pi_\phi(a^\prime|s^\prime)\right],
\end{equation}
where $\theta_i^\prime,i\in\{1,2\}$ are the parameters of the target networks, $\alpha\in\mathbb{R}_+$, and $a^\prime\sim\pi_\phi(\cdot|s^\prime)$. The way PAR uses to update its policy depends on the condition of the source domain, \emph{e.g.}, it can be either online or offline. We consider both conditions and discuss them below.

\textbf{\emph{\underline{Online PAR}}.} If the source domain is online, then the policy objective function gives:
\begin{equation}
    \label{eq:onlineactorloss}
    \mathcal{L}_{\rm actor}^{\rm on} = \mathbb{E}_{\substack{s\sim D_{\rm src}\cup D_{\rm tar}\\ a\sim\pi_\phi(\cdot|s)}}\left[ \min_{i=1,2} Q_{\theta_i}(s,a) - \alpha\log\pi_\phi(\cdot|s) \right].
\end{equation}

\textbf{\emph{\underline{Offline PAR}}.} Given an offline source domain, the deviation between the learned policy and the source domain behavior policy $\pi_{D_{\rm src}}$ ought to be considered based on Theorem \ref{theo:offline}. We then incorporate a behavior cloning term into the objective function of the policy, similar to \citet{Fujimoto2021AMA}. This term injects conservatism into policy learning on a fixed dataset and is necessary to mitigate the extrapolation error \cite{Fujimoto2019OffPolicyDR}, a challenge that is widely studied in offline RL \cite{Levine2020OfflineRL,Kumar2020ConservativeQF,lyu2022mildly,lyu2022state,kostrikov2022offline}. The policy objective function then yields:
\begin{equation}
\label{eq:offlineactorloss}
    \begin{split}
        \mathcal{L}_{\rm actor}^{\rm off} &= \mathbb{E}_{\substack{(s,a)\sim D_{\rm src}\\ \tilde{a}\sim\pi_\phi(\cdot|s)}} \left[ - (a - \tilde{a})^2 \right] + \lambda \times \mathcal{L}_{\rm actor}^{\rm on},
    \end{split}
\end{equation}
where $\lambda = \nicefrac{\nu}{\frac{1}{N}\sum_{(s_j,a_j)} \min_{i=1,2} Q_{\theta_i}(s_j,a_j)}$ is the normalization term that balances behavior cloning and maximizing the value function, $\nu\in\mathbb{R}_+$ is a hyperparameter, $\mathcal{L}_{\rm actor}^{\rm on}$ is the policy objective of online PAR in Equation \ref{eq:onlineactorloss}. The behavior cloning term ensures that the learned policy is close to the data-collecting policy of the source domain dataset. We summarize in Algorithm \ref{alg:parabstracted} the abstracted pseudocode of PAR, and defer the full pseudocodes to Appendix \ref{sec:pseudocodes}.

\begin{algorithm}[tb]
\caption{PAR (Abstracted Version)}
\label{alg:parabstracted}
{\bf Input:} Source domain $\mathcal{M}_{\rm src}$, target domain $\mathcal{M}_{\rm tar}$, target domain interaction interval $F$, batch size $N$

\begin{algorithmic}[1]
\STATE Initialize policy $\pi_\phi$, value functions $\{Q_{\theta_i}\}_{i=1,2}$ and target networks $\{Q_{\theta_i^\prime}\}_{i=1,2}$, replay buffers $\{D_{\rm src}, D_{\rm tar}\}$
\FOR{$i$ = 1, 2, ...}
\STATE (\emph{online}) Collect $(s_{\rm src},a_{\rm src},r_{\rm src},s_{\rm src}^\prime)$ in $\mathcal{M}_{\rm src}$ and store it, $D_{\rm src}\leftarrow D_{\rm src}\cup \{(s_{\rm src},a_{\rm src},r_{\rm src},s_{\rm src}^\prime)\}$
\IF{$i$\% $F$ == 0}
\STATE Interact with $\mathcal{M}_{\rm tar}$ and get $(s_{\rm tar},a_{\rm tar},r_{\rm tar},s_{\rm tar}^\prime)$. $D_{\rm tar}\leftarrow D_{\rm tar}\cup \{(s_{\rm tar},a_{\rm tar},r_{\rm tar},s_{\rm tar}^\prime)\}$
\ENDIF
\STATE Sample $N$ transitions from $D_{\rm tar}$
\STATE Train encoders in the target domain via Equation \ref{eq:representationobjective}
\STATE Sample $N$ transitions from $D_{\rm src}$
\STATE Modify source domain rewards with Equation \ref{eq:rewardmodification}
\STATE Update critics by minimizing Equation \ref{eq:criticloss}
\STATE (\emph{online}) Update actor by maximizing Equation \ref{eq:onlineactorloss}
\STATE (\emph{offline}) Update actor by maximizing Equation \ref{eq:offlineactorloss}
\STATE Update target networks
\ENDFOR
\end{algorithmic}
\end{algorithm}

\section{Experiments}
\label{sec:experiments}

In this section, we examine the effectiveness of our proposed method by conducting experiments on environments with kinematic and morphology discrepancies. We also extensively investigate the performance of our method under the offline source domain and different qualities of the offline datasets. Moreover, we empirically analyze the influence of the important hyperparameters in PAR.

\begin{figure*}
    \centering
    \includegraphics[width=0.99\linewidth]{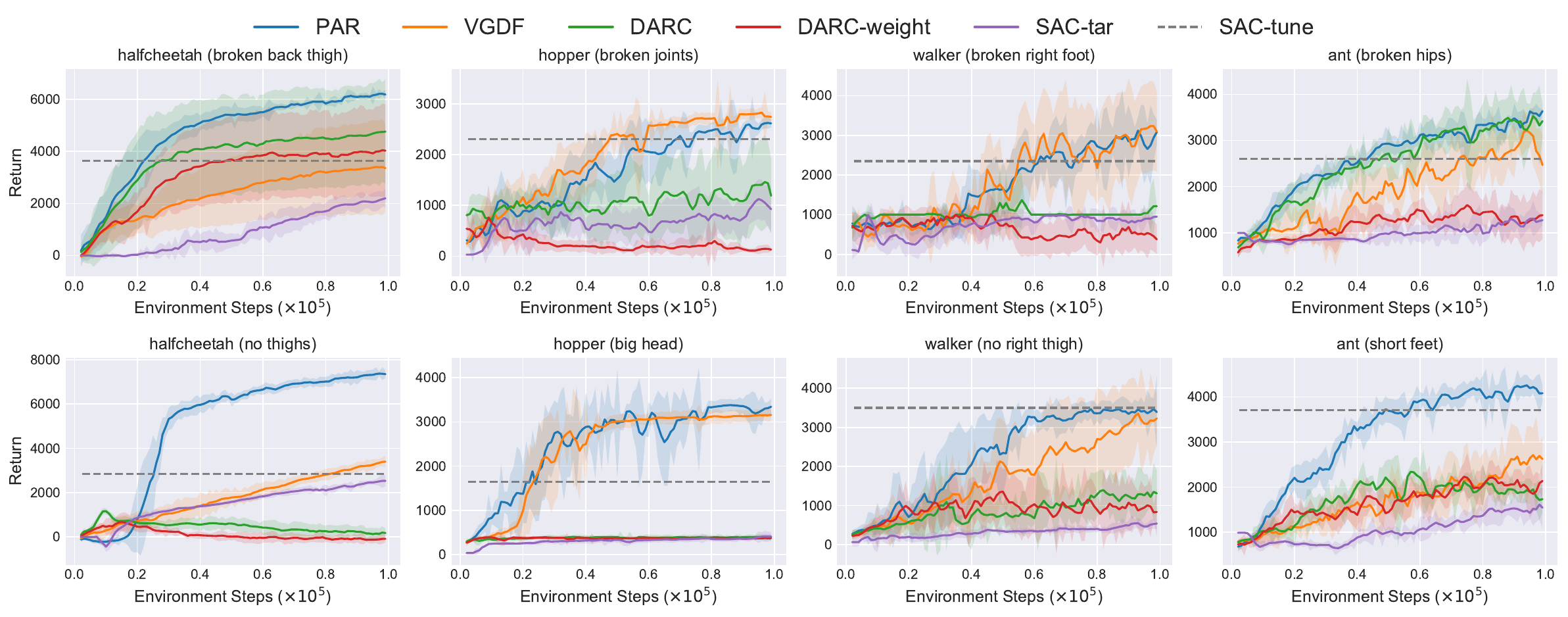}
    \caption{\textbf{Adaptation performance comparison when the source domain is online.} The curves depict the test performance of each algorithm \emph{in the target domain} under kinematic shifts (\emph{top}) and morphology shifts (\emph{bottom}). The modification to the environment is specified in the parentheses of the task name. The solid lines are the average returns over 5 different random seeds and the shaded region captures the standard deviation. The dashed line of SAC-tune denotes its final performance after fine-tuning $10^5$ steps.}
    \label{fig:onlineresult}
\end{figure*}

\subsection{Results with Online Source Domain}
\label{sec:onlineresults}
For the empirical evaluation of policy adaptation capabilities, we use four environments (\emph{halfcheetah}, \emph{hopper}, \emph{walker}, \emph{ant}) from OpenAI Gym \cite{Brockman2016OpenAIG} as source domains and modify their dynamics following \cite{Xu2023CrossDomainPA} to serve as target domains. The modifications include kinematic and morphology shifts, where we simulate broken joints of the robot by limiting the rotation angle of its joints (\emph{i.e.}, kinematic shifts), and we clip the size of some limbs of the simulated robot to realize morphology shifts. Please see details of the environment setting in Appendix \ref{sec:environmentsetting}.

We compare PAR against the following baselines: \textbf{SAC-tar} \cite{haarnoja2018softactorcritic}, which trains the SAC agent merely in the target domain for $10^5$ environmental steps; \textbf{DARC} \cite{eysenbach2021offdynamics}, which trains domain classifiers to estimate the dynamics discrepancy and leverage it to correct source domain rewards; \textbf{DARC-weight}, a variant of DARC that adopts the dynamics discrepancy term as importance sampling weights when updating critics; \textbf{VGDF} \cite{Xu2023CrossDomainPA}, a recent state-of-the-art method that filters transitions in the source domain that share similar value estimates as those in the target domain; \textbf{SAC-tune}, which trains the SAC agent in the source domain for 1M steps and fine-tunes it in the target domain with $10^5$ transitions. For online experiments, we allow all algorithms to interact 1M environmental steps with the source domain, but only $10^5$ steps in the target domain (\emph{i.e.}, the target domain interaction interval $F=10$). All algorithms are run with five random seeds. We defer implementation details to Appendix \ref{sec:implementationdetails}.

We summarize the comparison results in Figure \ref{fig:onlineresult}. Note that the evaluated environments are quite challenging, and baselines like DARC struggle for a good performance. Based on the curves, PAR outperforms SAC-tar on all of the tasks, indicating that our method successfully boosts the performance of the agent in the target domain by extracting useful knowledge from sufficient source domain data. Notably, PAR achieves the best performance on \textbf{6} out of 8 tasks, often surpassing baselines by a large margin. On the rest of the two tasks, PAR is able to achieve competitive performance against VGDF. PAR achieves \textbf{2x} sample efficiency compared to the best baseline method on tasks like \emph{halfcheetah (no thighs)}, \emph{ant (short feet)}, etc. Furthermore, PAR beats the fine-tuning method SAC-tune on \textbf{7} out of 8 tasks. These altogether illustrate the advantages of our method.

\subsection{Evaluations under Offline Source Domain}

There exist some circumstances where no real-time interaction with the source domain is available, but we have a previously gathered source domain dataset. We then investigate how our method behaves under this setting, and how the quality of the dataset affects the performance. To that end, we adopt datasets of the four environments (\emph{halfcheetah}, \emph{hopper}, \emph{walker}, \emph{ant}) from D4RL \cite{Fu2020D4RLDF} ``-v2" datasets with three quality levels (medium, medium-replay, medium-expert). This induces a total of \textbf{24} tasks.

We consider four baselines for comparison: \textbf{CQL-0} \cite{Kumar2020ConservativeQF}, which trains a CQL agent solely in the source offline dataset and then directly deploys the learned policy in the target domain in a zero-shot manner; \textbf{CQL+SAC}, which updates the offline source domain data with the CQL loss function, while the online target domain data with the SAC loss; \textbf{H2O} \cite{niu2022when}, which trains domain classifiers to estimate the dynamics gap and use it as an importance sampling weight for the bellman error of data from the source domain dataset; \textbf{VGDF+BC} \cite{Xu2023CrossDomainPA}, which incorporates an additional behavior cloning term in vanilla VGDF, similar to PAR. All algorithms have a limited budget of $10^5$ interactions with the target domain. The implementation details can be found in Appendix \ref{sec:implementationdetails}.

\begin{table*}[htb]
  \caption{\textbf{Performance comparison when the source domain is offline}, \emph{i.e.}, only static source domain datasets are available. We report the mean return in conjunction with standard deviation in the target domain under different dataset qualities of the source domain data (medium, medium-replay, medium-expert). The results are averaged over 5 varied random seeds. We \textbf{bold} and highlight the best cell.}
  \label{tab:offlineresults}
  \vspace{2mm}
  \centering
  \footnotesize
  \begin{tabular}{ll|llllllllll}
    \toprule
    Dataset Type  & Task Name & CQL-0 & CQL+SAC & H2O & VGDF+BC & PAR (ours) \\
    \midrule
    medium & halfcheetah (broken back thigh) & 1128$\pm$156 & 3967$\pm$204 & 5450$\pm$194 & 4834$\pm$250 & \cellcolor{green!20} \textbf{5686}$\pm$603 \\
    medium & halfcheetah (no thighs) & 361$\pm$29 & 1184$\pm$211 & 2863$\pm$209 & 3910$\pm$160 & \cellcolor{green!20} \textbf{5768}$\pm$117 \\
    medium & hopper (broken joints) & 155$\pm$19 & 498$\pm$73 & 2467$\pm$323 & 2785$\pm$75 & \cellcolor{green!20} \textbf{2825}$\pm$112 \\
    medium & hopper (big head) & 399$\pm$5 & 496$\pm$53 & 1451$\pm$480 & \cellcolor{green!20} \textbf{3060}$\pm$60 & 1450$\pm$143 \\
    medium & walker (broken right foot) & 1453$\pm$412 & 1877$\pm$1040 & 3309$\pm$418 & 3000$\pm$388 & \cellcolor{green!20} \textbf{3683}$\pm$211 \\
    medium & walker (no right thigh) & 975$\pm$131 & 1262$\pm$363 & 2225$\pm$546 & \cellcolor{green!20} \textbf{3293}$\pm$306 & 2899$\pm$841 \\
    medium & ant (broken hips) & 1230$\pm$99 & -1814$\pm$431 & 2704$\pm$253 & 1713$\pm$366 & \cellcolor{green!20} \textbf{3324}$\pm$72 \\
    medium & ant (short feet) & 1839$\pm$137 & -807$\pm$255 & 3892$\pm$85 & 3120$\pm$469 & \cellcolor{green!20} \textbf{4886}$\pm$97 \\
    \midrule
    medium-replay & halfcheetah (broken back thigh) & 655$\pm$226 & 3868$\pm$295 & 5103$\pm$35 & \cellcolor{green!20} \textbf{5398}$\pm$360 & 5227$\pm$445 \\
    medium-replay & halfcheetah (no thighs) & 398$\pm$63 & 575$\pm$619 & 3225$\pm$66 & 4271$\pm$162 & \cellcolor{green!20} \textbf{5161}$\pm$46 \\
    medium-replay & hopper (broken joints) & 1018$\pm$6 & 686$\pm$60 & 2325$\pm$193 & 2242$\pm$1057 & \cellcolor{green!20} \textbf{2376}$\pm$777 \\
    medium-replay & hopper (big head) & 365$\pm$7 & 556$\pm$222 & \cellcolor{green!20} \textbf{1854}$\pm$647 & 566$\pm$90 & 1336$\pm$419 \\
    medium-replay & walker (broken right foot) & 156$\pm$175 & 1018$\pm$22 & \cellcolor{green!20} \textbf{3536}$\pm$431 & 2901$\pm$1101 & 3128$\pm$1084 \\
    medium-replay & walker (no right thigh) & 337$\pm$189 & 1465$\pm$696 & \cellcolor{green!20} \textbf{4254}$\pm$207 & 2057$\pm$921 & 1249$\pm$706 \\
    medium-replay & ant (broken hips) & 882$\pm$28 & -1609$\pm$425 & 2497$\pm$190 & 2437$\pm$286 & \cellcolor{green!20} \textbf{2977}$\pm$186 \\
    medium-replay & ant (short feet) & 1294$\pm$191 & -1369$\pm$476 & 3782$\pm$382 & 4493$\pm$82 & \cellcolor{green!20} \textbf{4791}$\pm$102 \\
    \midrule
    medium-expert & halfcheetah (broken back thigh) & 843$\pm$510 & \cellcolor{green!20} \textbf{4283}$\pm$180 & 4100$\pm$211 & 3580$\pm$1801 & 3741$\pm$378 \\
    medium-expert & halfcheetah (no thighs) & 322$\pm$81 & 1669$\pm$439 & 1938$\pm$473 & 2740$\pm$297 & \cellcolor{green!20} \textbf{10517}$\pm$476 \\
    medium-expert & hopper (broken joints) & 458$\pm$441 & 1147$\pm$595 & 2587$\pm$252 & 2144$\pm$938 & \cellcolor{green!20} \textbf{2838}$\pm$339 \\
    medium-expert & hopper (big head) & 460$\pm$50 & 547$\pm$96 & 1156$\pm$574 & 2155$\pm$1182 & \cellcolor{green!20} \textbf{2676}$\pm$585 \\
    medium-expert & walker (broken right foot) & 813$\pm$459 & 2431$\pm$782 & 2254$\pm$710 & 1540$\pm$926 & \cellcolor{green!20} \textbf{4211}$\pm$196 \\
    medium-expert & walker (no right thigh) & 698$\pm$194 & 1547$\pm$346 & 2835$\pm$826 & 2047$\pm$1100 & \cellcolor{green!20} \textbf{4006}$\pm$1070 \\
    medium-expert & ant (broken hips) & 321$\pm$373 & 304$\pm$1458 & 2178$\pm$799 & 1868$\pm$321 & \cellcolor{green!20} \textbf{3113}$\pm$501 \\
    medium-expert & ant (short feet) & 1816$\pm$224 & -812$\pm$105 & 3511$\pm$441 & 1821$\pm$516 & \cellcolor{green!20} \textbf{4902}$\pm$34 \\
    \bottomrule
  \end{tabular}
\end{table*}

We present the comparison results in Table \ref{tab:offlineresults}. We observe that PAR also achieves superior performance given offline source domain datasets, surpassing baseline methods on \textbf{17} out of 24 tasks. It is worth mentioning that PAR is the only method that obtains meaningful performance on \emph{halfcheetah (no thighs)} with medium-expert dataset, which is approximately \textbf{4x} the performance of the strongest baseline. PAR is also the only method that can generally gain better performance on many tasks with higher quality datasets, \emph{e.g.}, despite that PAR has unsatisfying performance on \emph{hopper (big head)} under medium-level source domain dataset, its performance given the medium-expert source domain dataset is good. Nevertheless, methods like VGDF and H2O have worse performance given medium-expert datasets compared to medium-replay or medium datasets. These collectively show the superiority of PAR and shed light on capturing representation mismatch for cross-domain policy adaptation.

\subsection{Parameter Study}

Now we investigate the influence of two critical hyperparameters in PAR, reward penalty coefficient $\beta$, and target domain interaction interval $F$. Considering the page limit, please check more experimental results in Appendix \ref{sec:extendedexperiments}.

\begin{figure}[!t]
    \centering
    \subfigure[Penalty coefficient $\beta$.]{
    \label{fig:penaltycoefficient}
    \includegraphics[width=0.95\linewidth]{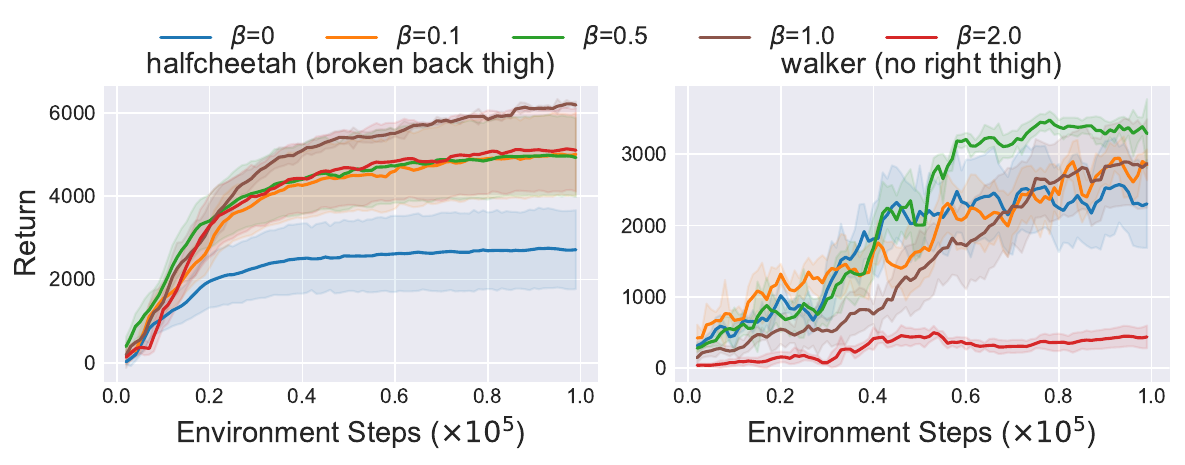}
    }\hspace{0mm}
    \subfigure[Target domain interaction interval $F$.]{
    \label{fig:targetinteractionfrequency}
    \includegraphics[width=0.95\linewidth]{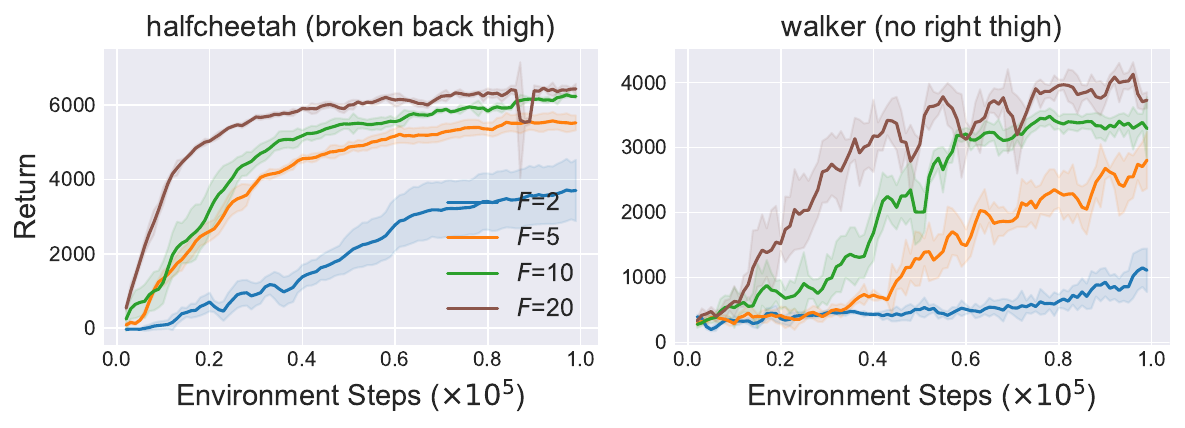}
    }
    \caption{\textbf{Parameter study of (a) reward penalty coefficient $\beta$, (b) target domain interaction interval $F$.} Results are averaged over 5 seeds and the shaded region denotes the standard deviation.}
    \label{fig:parameterstudycoefficientneighbor}
\end{figure}

\textbf{Penalty coefficient $\beta$.} $\beta$ controls the scale of the measured representation mismatch. Intuitively, the agent will struggle for good performance if $\beta$ is too large, and may fail to distinguish source domain samples with inconsistent dynamics if $\beta$ is too small. To examine its impact, we conduct experiments on two tasks with online source domains, \emph{halfcheetah (broken back thigh)} and \emph{walker (no right thigh)}. We evaluate PAR across $\beta\in\{0, 0.1, 0.5, 1.0, 2.0\}$, and show the results in Figure \ref{fig:penaltycoefficient}. We find that setting $\beta=0$ (\emph{i.e.}, no representation mismatch penalty) usually incurs a worse final performance, especially on the \emph{halfcheetah} task, verifying the necessity of the reward modification term. Figure \ref{fig:penaltycoefficient} also illustrates that the optimal $\beta$ can be task-dependent. We believe this is because different tasks have distinct inherent structures like rewards and state spaces. PAR exhibits some robustness to $\beta$, despite that employing a large $\beta$ may incur a performance drop on some tasks, \emph{e.g.}, on \emph{walker} task.

\textbf{Target domain interaction interval $F$.} $F$ decides how frequently the agent interacts with the target domain. Following Section \ref{sec:onlineresults}, only $10^5$ interactions with the target domain are permitted. We employ $F\in\{2,5,10,20\}$, and summarize the results in Figure \ref{fig:targetinteractionfrequency}, which show that PAR generally benefits from more source domain data incurred by a larger $F$, indicating that PAR can exploit dynamics-consistent transitions and realize efficient policy adaptation to another domain. We simply use $F=10$ by default.

\subsection{Runtime Comparison}

\begin{figure}
    \centering
    \includegraphics[width=0.7\linewidth]{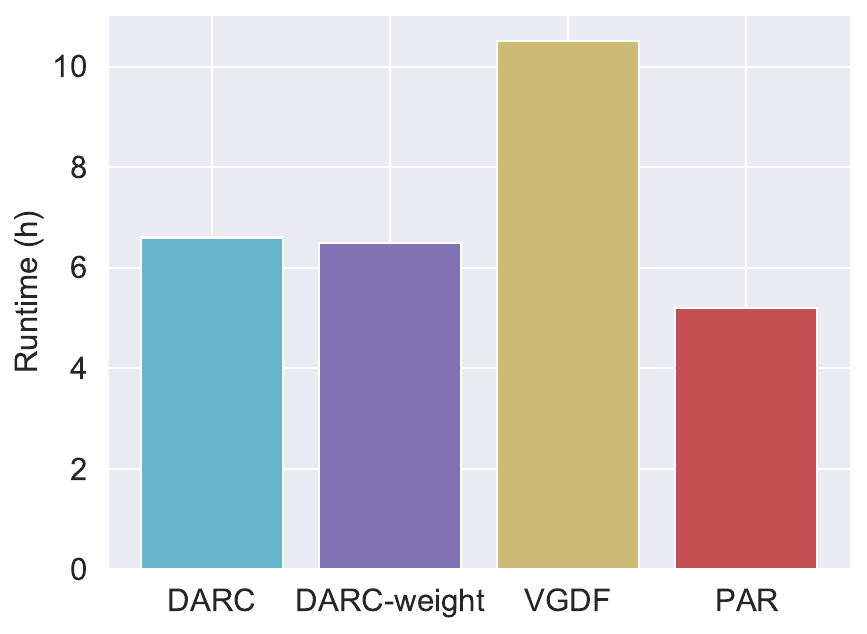}
    \caption{Runtime comparison of different methods.}
    \label{fig:runtime}
\end{figure}

Furthermore, we compare the runtime of PAR against baselines. All methods are run on the \emph{halfcheetah (broken back thigh)} task on a single GPU. The results in Figure \ref{fig:runtime} show that PAR is highly efficient in runtime thanks to training in the latent space with one state encoder and one state-action encoder. DARC and its variant have slightly larger training costs. VGDF consumes the most training time because it trains an ensemble of dynamics models in the original state space following model-based RL \cite{Janner2019WhenTT}.

\section{Discussions}
\label{sec:discussion}

In this section, we provide discussions on whether the performance of PAR can be largely affected if we use another representation learning objective, and why PAR beats DARC. We also explain the validity of the assumption. We believe these make a better understanding of our method.

\subsection{PAR with a Different Objective}

\begin{figure}[!htb]
    \centering
    \includegraphics[width=0.85\linewidth]{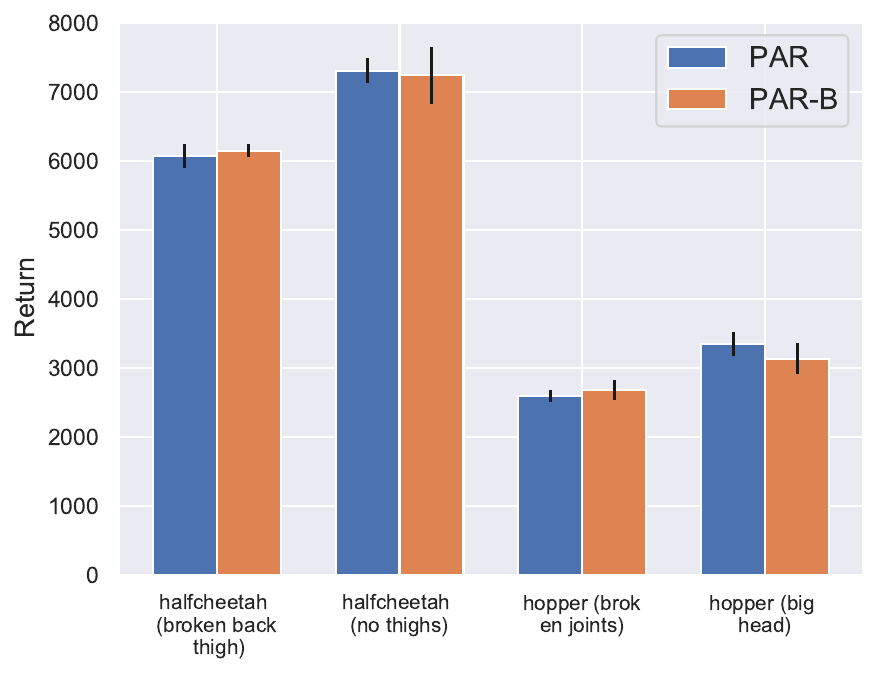}
    \caption{Performance comparison between PAR and PAR-B.}
    \label{fig:anotherobjective}
\end{figure}

We investigate how PAR behaves with a varying representation learning objective against Equation \ref{eq:representationobjective}. Such an objective needs to learn the latent dynamics information as well. To that end, we consider the following objective where $g$ now receives true state $s$ (instead of its representation) and action $a$ as inputs, and no stop gradient operator is required:
\begin{equation}
    \label{eq:residualencoderloss}
    \mathcal{L}^\prime(\psi, \xi) = \mathbb{E}_{(s,a,s^\prime)\sim D}\left[(g_{\xi}(s,a) - f_\psi(s^\prime) )^2 \right].
\end{equation}
Importantly, both $f$ and $g$ are optimized with this objective. Equation \ref{eq:residualencoderloss} also guarantees latent dynamics consistency. We tag this variant as PAR-B. To see how PAR-B competes against vanilla PAR, we conduct experiments on four tasks with kinematic and morphology mismatch. We report their final mean performance in the target domain in Figure \ref{fig:anotherobjective}, where only marginal return difference is observed between PAR and PAR-B, implying that another objective can also be valid as long as it embeds latent dynamics information.


\subsection{Why PAR Outperforms DARC?}

It is vital to address why PAR significantly outperforms DARC on numerous online tasks given that DARC also corrects source domain rewards (check Figure \ref{fig:onlineresult}). We stress that DARC learns \emph{domain classifiers} by leveraging both source domain data and target domain data and estimates the dynamics gap, which can be interpreted as how likely the measured source domain transition belongs to the target domain. However, if the transition deviates far from the target domain, the estimated gap $\log\frac{P_{\mathcal{M}_{\rm tar}}(s^\prime|s,a)}{P_{\mathcal{M}_{\rm src}}(s^\prime|s,a)}$ can be large and negatively affect the policy learning, which is similar in spirit to DARC's overly pessimistic issue that is criticized by \citet{Xu2023CrossDomainPA}. PAR, instead, captures representation mismatch by training encoders only with target domain data and evaluating representation deviations upon source domain data. We claim that PAR produces more appropriate reward penalties.

\begin{figure}[!htb]
    \centering
    \includegraphics[width=0.95\linewidth]{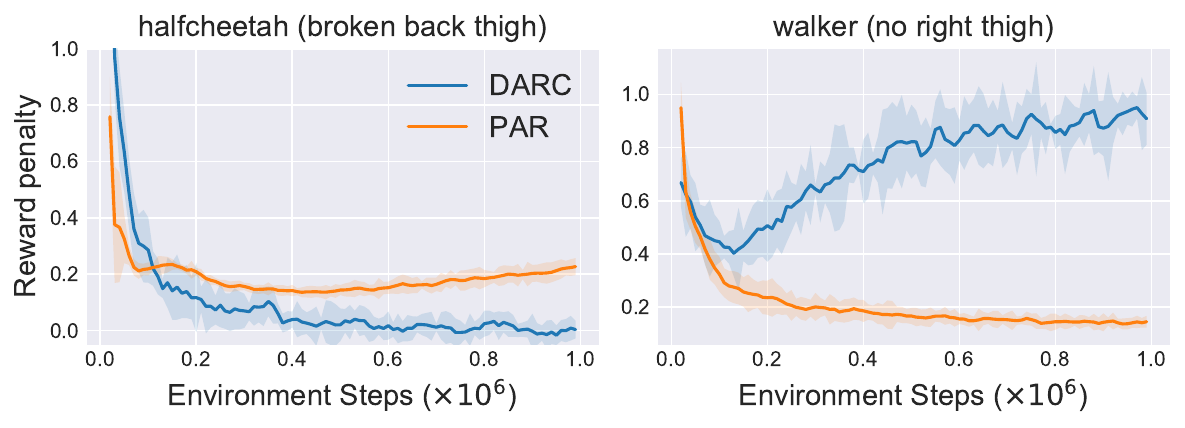}
    \caption{\textbf{Reward penalty comparison between DARC and PAR.} We record the average reward penalty across 5 seeds when training each method. The shaded region denotes the standard deviation.}
    \label{fig:darcparrewardcompare}
\end{figure}

To verify our claim, we log the reward penalties calculated by DARC and PAR, and summarize the results in Figure \ref{fig:darcparrewardcompare}. The reward penalty of PAR is large at first, while it decreases with more interactions, meaning that PAR uncovers more dynamics-consistent samples from the source domain. Note that the penalty by PAR tends to converge to a small number (not 0). However, the penalty from DARC is inconsistent on two tasks, \emph{i.e.}, it approaches 0 on \emph{halfcheetah} task while becomes large on \emph{walker} task. The results clearly indicate that capturing representation mismatch is a better choice.

\subsection{On the Rationality of the Assumption}
\label{sec:rationalityassumption}

In Assumption \ref{ass:1}, we assume a one-to-one mapping between $\mathcal{S}\times\mathcal{A}$ and $\mathcal{Z}$. One-to-one mapping mathematically indicates that the mapping is \emph{injective} (not necessarily surjective). That is, we only require there exists one unique $z$ in the latent space corresponding to the specific $(s,a)$ tuple. To satisfy this assumption, we first employ a deterministic state encoder $f$ and state-action encoder $g$ for representation learning, i.e., $f$ constructs a deterministic mapping from $\mathcal{S}$ to $\mathcal{Z}$ and $g$ is a deterministic mapping from $\mathcal{S}\times\mathcal{A}$ to $\mathcal{Z}$. It remains to decide whether the mapped representation is unique. Note that it is the user's choice of what representation learning approach and which latent representation space to use. One can surely choose a representation method and representation space to let the assumption hold. With our adopted representation learning formula in Equation \ref{eq:representationobjective}, it is less likely that two distinct $(s,a)$ tuples are mapped into the same latent vector because that indicates they share the same dynamics transition information (since Equation \ref{eq:representationobjective} realizes latent dynamics consistency). To further mitigate this concern, the dimension of the state-action representation in PAR is much larger (it is set to be $256$ as shown in Table \ref{tab:hyperparametersetup} in the appendix) than the input state vector and action vector. We believe these explain the rationality of the assumption.

\section{Conclusion and Limitations}
\label{sec:conclusion}

In this paper, we study how to effectively adapt policies to another domain with dynamics discrepancy. We propose a novel algorithm, Policy Adaptation by Representation Mismatch (PAR), which captures the representation mismatch between the source domain and the target domain, and employs the resulting representation deviation to compensate source domain rewards. Our method is motivated and supported by rigorous theoretical analysis. Experimental results demonstrate that PAR achieves strong performance and outperforms recent strong baselines under scenarios like kinematic shifts and morphology mismatch, regardless of whether the source domain is online or offline.

Despite the effectiveness of our method, we have to admit that there exist some limitations of our work. First, one may need to decide the best $\beta$ manually in practice. Second, PAR behaves less satisfyingly given some (not all of them) medium-replay source domain datasets, suggesting that it may be hard for PAR to handle datasets with large diversities. For future work, it is interesting to design mechanisms to adaptively tune $\beta$, and enable PAR to consistently acquire a good performance provided datasets with large diversities.

\section*{Acknowledgements}

This work was supported by the STI 2030-Major Projects under Grant 2021ZD0201404 and the NSFC under Grant 62250068. This work was done when Jiafei Lyu worked as an intern at Tencent IEG. The authors thank Liangpeng Zhang for providing advice on the draft of this work. The authors also would like to thank the anonymous reviewers for their valuable comments on our manuscript.

\section*{Impact Statement}

This paper presents work whose goal is to advance the field of 
Machine Learning. There are many potential societal consequences 
of our work, none which we feel must be specifically highlighted here.

\nocite{langley00}

\bibliography{example_paper}
\bibliographystyle{icml2024}

\newpage
\appendix
\onecolumn
\section{Missing Proofs}
\label{sec:missingproofs}

In this section, we formally present all the missing proofs from the main text. For better readability, we restate theorems in the appendix. We also need some lemmas, which can be found in Appendix \ref{sec:usefullemma}.

\subsection{Proof of Theorem \ref{theo:mutual}}

\begin{theorem}
\label{apptheo:mutual}
    For any $(s,a)$, denote its representation as $z$, and suppose $s_{\rm src}^\prime\sim P_{\mathcal{M}_{\rm src}}(\,\cdot\,|s,a)$, $s_{\rm tar}^\prime\sim P_{\mathcal{M}_{\rm tar}}(\,\cdot\,|s,a)$. Denote $h(z;s_{\rm src}^\prime,s_{\rm tar}^\prime)=I(z;s_{\rm tar}^\prime) - I(z;s_{\rm src}^\prime)$, then we have measuring $h(z;s_{\rm src}^\prime,s_{\rm tar}^\prime)$ is equivalent to measuring  the representation deviation $D_{\rm KL}(P(z|s_{\rm tar}^\prime) \| P(z|s_{\rm src}^\prime))$.
\end{theorem}

\begin{proof}
    By the definition of mutual information, we have
    \begin{align*}
        h(z;s_{\rm src}^\prime,s_{\rm tar}^\prime) &= I(z;s_{\rm tar}^\prime) - I(z;s_{\rm src}^\prime) \\
        &= \int_{\mathcal{Z}} \int_{\mathcal{S}} P(z,s_{\rm tar}^\prime)\log\dfrac{P(z,s_{\rm tar}^\prime)}{P(z)P(s_{\rm tar}^\prime)}dz ds_{\rm tar}^\prime - \int_{\mathcal{Z}} \int_{\mathcal{S}} P(z,s_{\rm src}^\prime)\log\dfrac{P(z,s_{\rm src}^\prime)}{P(z)P(s_{\rm src}^\prime)}dz ds_{\rm src}^\prime \\
        &= \int_{\mathcal{Z}} \int_{\mathcal{S}} P(z,s_{\rm tar}^\prime)\log\dfrac{P(z|s_{\rm tar}^\prime)}{P(z)}dz ds_{\rm tar}^\prime - \int_{\mathcal{Z}} \int_{\mathcal{S}} P(z,s_{\rm src}^\prime)\log\dfrac{P(z|s_{\rm src}^\prime)}{P(z)}dz ds_{\rm src}^\prime \\
        &= \int_{\mathcal{Z}} \int_{\mathcal{S}} \int_{\mathcal{S}} P(z,s_{\rm tar}^\prime,s_{\rm src}^\prime)\log\dfrac{P(z|s_{\rm tar}^\prime)}{P(z)}dz ds_{\rm tar}^\prime ds_{\rm src}^\prime - \int_{\mathcal{Z}} \int_{\mathcal{S}} \int_{\mathcal{S}} P(z,s_{\rm src}^\prime,s_{\rm tar}^\prime)\log\dfrac{P(z|s_{\rm src}^\prime)}{P(z)}dz ds_{\rm src}^\prime ds_{\rm tar}^\prime \\
        &= \int_{\mathcal{Z}} \int_{\mathcal{S}} \int_{\mathcal{S}} P(z,s_{\rm tar}^\prime,s_{\rm src}^\prime)\log\dfrac{P(z|s_{\rm tar}^\prime)}{P(z|s_{\rm src}^\prime)}dz ds_{\rm tar}^\prime ds_{\rm src}^\prime \\
        &= D_{\rm KL}\left( P(z|s_{\rm tar}^\prime) \| P(z|s_{\rm src}^\prime) \right). \qquad\qquad\quad (\rm By\, the \, definition\, of \, Kullback–Leibler\, divergence)
    \end{align*}
    We then can conclude that measuring the defined function $h(z;s_{\rm src}^\prime,s_{\rm tar}^\prime)$ is equivalent to measuring the KL-divergence between $P(z|s_{\rm tar}^\prime)$ and $P(z|s_{\rm src}^\prime)$, which is the deviation of representations given the source domain state and target domain state, respectively. Note that the definition of the KL-divergence already involves expectations over $s_{\rm src}^\prime$ and $s_{\rm tar}^\prime$. While one can also write $\mathbb{E}_{s_{\rm src}^\prime,s_{\rm tar}^\prime}[D_{\rm KL}\left( P(z|s_{\rm tar}^\prime) \| P(z|s_{\rm src}^\prime) \right)]$ and it should not affect the result.
\end{proof}

\subsection{Proof of Theorem \ref{theo:equal}}

\begin{theorem}
\label{apptheo:equal}
     Measuring the representation deviation between the source domain and the target domain is equivalent to measuring the dynamics mismatch between two domains. Formally, we can derive that $D_{\rm KL}\left( P(z|s_{\rm tar}^\prime) \| P(z|s_{\rm src}^\prime) \right) = D_{\rm KL}\left( P(s_{\rm tar}^\prime|z) \| P(s_{\rm src}^\prime|z) \right) + \mathbb{H}(s_{\rm tar}^\prime) - \mathbb{H}(s_{\rm src}^\prime).$
\end{theorem}

\begin{proof}
    We would like to establish a connection between the representation deviations in the two domains and the dynamics discrepancies between the two domains. We achieve this by rewriting the defined function $h(z;s_{\rm src}^\prime,s_{\rm tar}^\prime)$ as follows,
    \begin{align*}
        h(z;s_{\rm src}^\prime,s_{\rm tar}^\prime) &= I(z;s_{\rm tar}^\prime) - I(z;s_{\rm src}^\prime) \\
        &= \int_{\mathcal{Z}} \int_{\mathcal{S}} P(z,s_{\rm tar}^\prime)\log\dfrac{P(z,s_{\rm tar}^\prime)}{P(z)P(s_{\rm tar}^\prime)}dz ds_{\rm tar}^\prime - \int_{\mathcal{Z}} \int_{\mathcal{S}} P(z,s_{\rm src}^\prime)\log\dfrac{P(z,s_{\rm src}^\prime)}{P(z)P(s_{\rm src}^\prime)}dz ds_{\rm src}^\prime \\
        &= \int_{\mathcal{Z}} \int_{\mathcal{S}} P(z,s_{\rm tar}^\prime)\log\dfrac{P(s_{\rm tar}^\prime|z)}{P(s_{\rm tar}^\prime)}dz ds_{\rm tar}^\prime - \int_{\mathcal{Z}} \int_{\mathcal{S}} P(z,s_{\rm src}^\prime)\log\dfrac{P(s_{\rm src}^\prime|z)}{P(s_{\rm src}^\prime)}dz ds_{\rm src}^\prime \\
        &= \int_{\mathcal{Z}} \int_{\mathcal{S}} \int_{\mathcal{S}} P(z,s_{\rm tar}^\prime,s_{\rm src}^\prime)\log\dfrac{P(s_{\rm tar}^\prime|z)}{P(s_{\rm tar}^\prime)}dz ds_{\rm tar}^\prime ds_{\rm src}^\prime - \int_{\mathcal{Z}} \int_{\mathcal{S}} \int_{\mathcal{S}} P(z,s_{\rm src}^\prime,s_{\rm tar}^\prime)\log\dfrac{P(s_{\rm src}^\prime|z)}{P(s_{\rm src}^\prime)}dz ds_{\rm src}^\prime ds_{\rm tar}^\prime \\
        &= \int_{\mathcal{Z}} \int_{\mathcal{S}} \int_{\mathcal{S}} P(z,s_{\rm tar}^\prime,s_{\rm src}^\prime)\log\dfrac{P(s_{\rm tar}^\prime|z)}{P(s_{\rm src}^\prime|z)}dz ds_{\rm tar}^\prime ds_{\rm src}^\prime - \int_{\mathcal{S}}P(s_{\rm tar}^\prime)\log P(s_{\rm tar}^\prime)ds_{\rm tar}^\prime \\
        & \qquad\qquad \qquad\qquad\qquad\qquad\qquad\qquad\qquad\qquad\qquad\qquad + \int_{\mathcal{S}}P(s_{\rm src}^\prime)\log P(s_{\rm src}^\prime)ds_{\rm src}^\prime \\
        &= D_{\rm KL}\left( P(s_{\rm tar}^\prime|z) \| P(s_{\rm src}^\prime|z) \right) + \mathbb{H}(s_{\rm tar}^\prime) - \mathbb{H}(s_{\rm src}^\prime).
    \end{align*}
    One can see that the defined function is also connected to the dynamics discrepancy term $D_{\rm KL}(P(s_{\rm tar}^\prime)|z)\|P(s_{\rm src}^\prime|z)$. It also correlates to two entropy terms. Nevertheless, we observe that the source domain and the target domain are specified and fixed, and their state distributions are also fixed, indicating that the entropy terms are constants. Then by using the conclusion from Theorem \ref{theo:mutual}, we have
    \begin{equation}
    \label{eq:klequalterm}
        \underbrace{D_{\rm KL}\left( P(z|s_{\rm tar}^\prime) \| P(z|s_{\rm src}^\prime) \right)}_{\rm representation \, deviation} = \underbrace{D_{\rm KL}\left( P(s_{\rm tar}^\prime|z) \| P(s_{\rm src}^\prime|z) \right)}_{\rm dynamics \, deviation} + \underbrace{\mathbb{H}(s_{\rm tar}^\prime) - \mathbb{H}(s_{\rm src}^\prime)}_{\rm constants}.
    \end{equation}
    Hence, we conclude that measuring representation deviations between two domains is equivalent to measuring the dynamics mismatch.
\end{proof}

\subsection{Proof of Theorem \ref{theo:online}}
\label{sec:prooftheo3}

\begin{theorem}[Online performance bound]
    \label{apptheo:online}
    Denote $\mathcal{M}_{\rm src}$, $\mathcal{M}_{\rm tar}$ as the source domain and the target domain, respectively, then the return difference of any policy $\pi$ between $\mathcal{M}_{\rm src}$ and $\mathcal{M}_{\rm tar}$ is bounded:
    \begin{align*}
        J_{\mathcal{M}_{\rm tar}}(\pi) - J_{\mathcal{M}_{\rm src}}(\pi) \ge - \dfrac{\sqrt{2}\gamma r_{\rm max}}{(1-\gamma)^2}\underbrace{\mathbb{E}_{\rho_{\mathcal{M}_{\rm src}}^\pi}\left[\sqrt{D_{\rm KL}(P(z|s_{\rm src}^\prime) \| P(z|s_{\rm tar}^\prime))}\right]}_{(a): \rm representation\, mismatch} - \dfrac{\sqrt{2}\gamma r_{\rm max}}{(1-\gamma)^2}  \underbrace{\mathbb{E}_{\rho_{\mathcal{M}_{\rm src}}^\pi} \left[\sqrt{|\mathbb{H}(s_{\rm src}^\prime) - \mathbb{H}(s_{\rm tar}^\prime)|}\right]}_{(b): \rm state\, distribution\, deviation}. 
    \end{align*}
\end{theorem}

\begin{proof}
    To show this theorem, we reiterate the Assumption \ref{ass:1} we made in the main text, \emph{i.e.}, the state-action pair $(s,a)$ and its corresponding representation $z$ are a one-to-one mapping from the original space $\mathcal{S}\times\mathcal{A}$ to the latent space $\mathcal{Z}$. This indicates that we could construct a pseudo probability distribution given the representation $z$ that is the same as the transition dynamics probability in the system, \emph{i.e.}, $P(s^\prime_{\rm src}|z) = P(s_{\rm src}^\prime|s,a)=P_{\mathcal{M}_{\rm src}}(\cdot|s,a), P(s^\prime_{\rm tar}|z) = P(s_{\rm tar}^\prime|s,a) = P_{\mathcal{M}_{\rm tar}}(\cdot|s,a), \forall\, s,a$.

    Recall that the value function $V(s)$ estimates the expected return given the state $s$, and state-action value function $Q(s,a)$ estimates the expected return given the state $s$ and action $a$. Since the rewards are bounded, we have $|V(s)|\le \dfrac{r_{\rm max}}{1-\gamma}, |Q(s,a)|\le \dfrac{r_{\rm max}}{1-\gamma},\forall\, s,a$. We denote value function under policy $\pi$ and MDP $\mathcal{M}$ as $V_{\mathcal{M}}^\pi(s),Q_{\mathcal{M}}^\pi(s,a)$, respectively.

    By using Lemma \ref{lemma:telescopr}, we have
    \begin{align*}
        J_{\mathcal{M}_{\rm src}}(\pi) - J_{\mathcal{M}_{\rm tar}}(\pi) &= \dfrac{\gamma}{1-\gamma}\mathbb{E}_{\rho_{\mathcal{M}_{\rm src}}^\pi(s,a)}\left[ \int_{s^\prime} P_{\mathcal{M}_{\rm src}}(s^\prime|s,a)V_{\mathcal{M}_{\rm tar}}^\pi(s^\prime)ds^\prime - \int_{s^\prime} P_{\mathcal{M}_{\rm tar}}(s^\prime|s,a)V_{\mathcal{M}_{\rm tar}}^\pi(s^\prime)ds^\prime \right] \\
        &=\dfrac{\gamma}{1-\gamma}\mathbb{E}_{\rho_{\mathcal{M}_{\rm src}}^\pi(s,a)}\left[ \int_{s^\prime} (P_{\mathcal{M}_{\rm src}}(s^\prime|s,a) - P_{\mathcal{M}_{\rm tar}}(s^\prime|s,a))V_{\mathcal{M}_{\rm tar}}^\pi(s^\prime)ds^\prime \right] \\
        &\le \dfrac{\gamma}{1-\gamma}\mathbb{E}_{\rho_{\mathcal{M}_{\rm src}}^\pi(s,a)}\left| \int_{s^\prime} (P_{\mathcal{M}_{\rm src}}(s^\prime|s,a) - P_{\mathcal{M}_{\rm tar}}(s^\prime|s,a))V_{\mathcal{M}_{\rm tar}}^\pi(s^\prime)ds^\prime \right| \\
        &\le \dfrac{\gamma}{1-\gamma}\mathbb{E}_{\rho_{\mathcal{M}_{\rm src}}^\pi(s,a)}\left[ \int_{s^\prime} \left|P_{\mathcal{M}_{\rm src}}(s^\prime|s,a) - P_{\mathcal{M}_{\rm tar}}(s^\prime|s,a)\right|\times \left|V_{\mathcal{M}_{\rm tar}}^\pi(s^\prime)\right|ds^\prime \right] \\
        &\le \dfrac{\gamma r_{\rm max}}{(1-\gamma)^2}\mathbb{E}_{\rho_{\mathcal{M}_{\rm src}}^\pi(s,a)}\left[ \int_{s^\prime} \left|P_{\mathcal{M}_{\rm src}}(s^\prime|s,a) - P_{\mathcal{M}_{\rm tar}}(s^\prime|s,a)\right| ds^\prime \right] \\
        &= \dfrac{2\gamma r_{\rm max}}{(1-\gamma)^2}\mathbb{E}_{\rho_{\mathcal{M}_{\rm src}}^\pi(s,a)}\left[ D_{\rm TV}\left( P_{\mathcal{M}_{\rm src}}(s^\prime|s,a)\| P_{\mathcal{M}_{\rm tar}}(s^\prime|s,a) \right) \right] \\
        &= \dfrac{2\gamma r_{\rm max}}{(1-\gamma)^2}\mathbb{E}_{\rho_{\mathcal{M}_{\rm src}}^\pi(s,a)}\left[ D_{\rm TV}\left( P(s_{\rm src}^\prime|z)\| P(s_{\rm tar}^\prime|z) \right) \right] \\
        &\le \dfrac{2\gamma r_{\rm max}}{(1-\gamma)^2}\mathbb{E}_{\rho_{\mathcal{M}_{\rm src}}^\pi(s,a)}\left[ \sqrt{\dfrac{1}{2} D_{\rm KL}\left( P(s_{\rm src}^\prime|z)\| P(s_{\rm tar}^\prime|z) \right)} \right] \qquad\qquad\qquad (i) \\
        &\le \dfrac{\sqrt{2}\gamma r_{\rm max}}{(1-\gamma)^2}\mathbb{E}_{\rho_{\mathcal{M}_{\rm src}}^\pi(s,a)}\left[ \sqrt{ D_{\rm KL}\left( P(z|s_{\rm src}^\prime)\| P(z|s_{\rm tar}^\prime) \right)} \right] \\
        &\qquad + \dfrac{\sqrt{2}\gamma r_{\rm max}}{(1-\gamma)^2} \mathbb{E}_{\rho_{\mathcal{M}_{\rm src}}^\pi(s,a)} \left[ \sqrt{ |\mathbb{H}(s_{\rm src}^\prime) - \mathbb{H}(s_{\rm tar}^\prime) } \right],
    \end{align*}
    where $D_{\rm TV}(p,q)$ denotes the total variation distance between two distribution $p$ and $q$, the inequality $(i)$ is due to the Pinsker's inequality \cite{csiszar2011information}, and the last step is by using Equation \ref{eq:klequalterm} and the triangle inequality. Then we conclude the proof.
\end{proof}

\subsection{Proof of Theorem \ref{theo:offline}}

\begin{theorem}[Offline performance bound]
\label{apptheo:offline}
    Denote the empirical policy distribution in the offline dataset $D$ from source domain $\mathcal{M}_{\rm src}$ as $\pi_D:=\frac{\sum_{D}\mathbbm{1}(s,a)}{\sum_{D}\mathbbm{1}(s)}$, then the return difference of any policy $\pi$ between the source domain $\mathcal{M}_{\rm src}$ and the target domain $\mathcal{M}_{\rm tar}$ is bounded:
    \begin{align*}
        J_{\mathcal{M}_{\rm tar}}(\pi) - J_{\mathcal{M}_{\rm src}}(\pi) \ge - \dfrac{4r_{\rm max}}{(1-\gamma)^2}\underbrace{\mathbb{E}_{\rho_{\mathcal{M}_{\rm src}}^{\pi_D},P_{\mathcal{M}_{\rm src}}}[D_{\rm TV}(\pi_D || \pi )]}_{(a): \rm policy\, deviation}
        &- \dfrac{\sqrt{2}\gamma r_{\rm max}}{(1-\gamma)^2}\underbrace{\mathbb{E}_{\rho_{\mathcal{M}_{\rm src}}^{\pi_D}} \left[\sqrt{D_{\rm KL}(P(z|s_{\rm src}^\prime) \| P(z|s_{\rm tar}^\prime))}\right]}_{(b): \rm representation\, mismatch} \\
        &- \dfrac{\sqrt{2}\gamma r_{\rm max}}{(1-\gamma)^2} \underbrace{\mathbb{E}_{\rho_{\mathcal{M}_{\rm src}}^{\pi_D}} \left[\sqrt{|\mathbb{H}(s_{\rm src}^\prime) - \mathbb{H}(s_{\rm tar}^\prime)|}\right]}_{(c): \rm state\, distribution\, deviation}. 
    \end{align*}
\end{theorem}

\begin{proof}
    Since it is infeasible to directly interact with the source domain, and we have the empirical policy distribution $\pi_D$ in the offline dataset, we bound the performance difference by involving the term $J_{\mathcal{M}_{\rm src}}(\pi_D)$ We have
    \begin{equation}
    \label{eq:bounddiv}
        J_{\mathcal{M}_{\rm tar}}(\pi) - J_{\mathcal{M}_{\rm src}}(\pi) = \underbrace{\left( J_{\mathcal{M}_{\rm tar}}(\pi) - J_{\mathcal{M}_{\rm src}}(\pi_D) \right)}_{(a)} + \underbrace{\left( J_{\mathcal{M}_{\rm src}}(\pi_D) - J_{\mathcal{M}_{\rm src}}(\pi) \right)}_{(b)}.
    \end{equation}
    The term $(a)$ depicts the performance of the learned policy in the target domain against the performance of the data-collecting policy in the offline dataset, and the term $(b)$ measures the performance deviation between the learned policy and the behavior policy in the source domain. We first bound term $(b)$. By using Lemma \ref{lemma:actionlemma}, we have
    \begin{align*}
        J_{\mathcal{M}_{\rm src}}(\pi_D) - J_{\mathcal{M}_{\rm src}}(\pi) &= \dfrac{1}{1-\gamma} \mathbb{E}_{\rho_{\mathcal{M}_{\rm src}}^{\pi_D}(s,a),s^\prime\sim P_{\mathcal{M}_{\rm src}}(\cdot|s,a)}\left[ \mathbb{E}_{a^\prime\sim\pi_D}\left[Q_{\mathcal{M}_{\rm src}}^\pi(s^\prime,a^\prime)\right] - \mathbb{E}_{a^\prime\sim\pi}\left[Q_{\mathcal{M}_{\rm src}}^\pi(s^\prime,a^\prime)\right] \right] \\
        & \ge -\dfrac{1}{1-\gamma} \mathbb{E}_{\rho_{\mathcal{M}_{\rm src}}^{\pi_D}(s,a),s^\prime\sim P_{\mathcal{M}_{\rm src}}(\cdot|s,a)}\left| \mathbb{E}_{a^\prime\sim\pi_D}\left[Q_{\mathcal{M}_{\rm src}}^\pi(s^\prime,a^\prime)\right] - \mathbb{E}_{a^\prime\sim\pi}\left[Q_{\mathcal{M}_{\rm src}}^\pi(s^\prime,a^\prime)\right] \right| \\
        & \ge -\dfrac{1}{1-\gamma} \mathbb{E}_{\rho_{\mathcal{M}_{\rm src}}^{\pi_D}(s,a),s^\prime\sim P_{\mathcal{M}_{\rm src}}(\cdot|s,a)}\left| \sum_{a^\prime\in \mathcal{A}}(\pi_D(a^\prime|s^\prime) - \pi(a^\prime|s^\prime)) Q_{\mathcal{M}_{\rm src}}^\pi(s^\prime,a^\prime) \right| \\
        &\ge -\dfrac{r_{\rm max}}{(1-\gamma)^2} \mathbb{E}_{\rho_{\mathcal{M}_{\rm src}}^{\pi_D}(s,a),s^\prime\sim P_{\mathcal{M}_{\rm src}}(\cdot|s,a)}\left| \sum_{a^\prime\in \mathcal{A}}(\pi_D(a^\prime|s^\prime) - \pi(a^\prime|s^\prime)) \right| \\
        &= -\dfrac{2r_{\rm max}}{(1-\gamma)^2} \mathbb{E}_{\rho_{\mathcal{M}_{\rm src}}^{\pi_D}(s,a),s^\prime\sim P_{\mathcal{M}_{\rm src}}(\cdot|s,a)} \left[ D_{\rm TV}(\pi_D(\cdot|s^\prime)\| \pi(\cdot|s^\prime) \right].
    \end{align*}
    It remains to bound term $(a)$. By using Lemma \ref{lemma:extendedtelescopr}, we have
    \begin{align*}
        J_{\mathcal{M}_{\rm tar}}(\pi) - J_{\mathcal{M}_{\rm src}}(\pi_D) &= - \dfrac{1}{1-\gamma}\mathbb{E}_{\rho_{\mathcal{M}_{\rm src}}^{\pi_D}(s,a)}\left[ \mathbb{E}_{s_{\rm src}^\prime\sim P_{\mathcal{M}_{\rm src}},a^\prime\sim\pi_D}[Q_{\mathcal{M}_{\rm tar}}^\pi(s^\prime_{\rm src},a^\prime)] - \mathbb{E}_{s_{\rm tar}^\prime\sim P_{\mathcal{M}_{\rm tar}},a^\prime\sim\pi}[Q_{\mathcal{M}_{\rm tar}}^\pi(s^\prime_{\rm tar},a^\prime)] \right] \\
        &= - \dfrac{1}{1-\gamma}\mathbb{E}_{\rho_{\mathcal{M}_{\rm src}}^{\pi_D}(s,a)}\left[ \underbrace{(\mathbb{E}_{s_{\rm src}^\prime\sim P_{\mathcal{M}_{\rm src}},a^\prime\sim\pi_D}[Q_{\mathcal{M}_{\rm tar}}^{\pi}(s^\prime_{\rm src},a^\prime)] - \mathbb{E}_{s_{\rm src}^\prime\sim P_{\mathcal{M}_{\rm src}},a^\prime\sim\pi}[Q_{\mathcal{M}_{\rm tar}}^{\pi}(s^\prime_{\rm src},a^\prime)])}_{(c)} \right. \\
        & \qquad\qquad\qquad \left. + \underbrace{(\mathbb{E}_{s_{\rm src}^\prime\sim P_{\mathcal{M}_{\rm src}},a^\prime\sim\pi}[Q_{\mathcal{M}_{\rm tar}}^{\pi}(s^\prime_{\rm src},a^\prime)] - \mathbb{E}_{s_{\rm tar}^\prime\sim P_{\mathcal{M}_{\rm tar}},a^\prime\sim\pi}[Q_{\mathcal{M}_{\rm tar}}^{\pi}(s^\prime_{\rm tar},a^\prime)])}_{(d)} \right].
    \end{align*}
    We bound term $(c)$ as follows:
    \begin{align*}
        (c) &= \mathbb{E}_{s_{\rm src}^\prime\sim P_{\mathcal{M}_{\rm src}}}\left[ \sum_{a^\prime\in\mathcal{A}} (\pi_D(a^\prime|s^\prime_{\rm src}) - \pi(a^\prime|s^\prime_{\rm src})) Q_{\mathcal{M}_{\rm tar}}^{\pi}(s^\prime_{\rm src},a^\prime)\right] \\
        &\le \mathbb{E}_{s_{\rm src}^\prime\sim P_{\mathcal{M}_{\rm src}}}\left[ \sum_{a^\prime\in\mathcal{A}} |\pi_D(a^\prime|s^\prime_{\rm src}) - \pi(a^\prime|s^\prime_{\rm src})|\times | Q_{\mathcal{M}_{\rm tar}}^{\pi}(s^\prime_{\rm src},a^\prime)|\right] \\
        &\le \dfrac{2r_{\rm max}}{1-\gamma}  \mathbb{E}_{s^\prime\sim P_{\mathcal{M}_{\rm src}}} \left[ D_{\rm TV}(\pi_D(\cdot|s^\prime)\| \pi(\cdot|s^\prime) \right].
    \end{align*}
    Finally, we bound term $(d)$.
    \begin{align*}
        (d) &= \mathbb{E}_{s^\prime\sim P_{\mathcal{M}_{\rm src}},a^\prime\sim\pi}[Q_{\mathcal{M}_{\rm tar}}^{\pi}(s^\prime,a^\prime)] - \mathbb{E}_{s^\prime\sim P_{\mathcal{M}_{\rm tar}},a^\prime\sim\pi}[Q_{\mathcal{M}_{\rm tar}}^{\pi}(s^\prime,a^\prime)] \\
        &= \mathbb{E}_{a^\prime\sim\pi}\left[ \int_\mathcal{S} (P_{\mathcal{M}_{\rm src}}(s^\prime|s,a) - P_{\mathcal{M}_{\rm tar}}(s^\prime|s,a)) Q_{\mathcal{M}_{\rm tar}}^\pi(s^\prime,a^\prime) ds^\prime \right] \\
        &\le \mathbb{E}_{a^\prime\sim\pi}\left[ \int_\mathcal{S} |P_{\mathcal{M}_{\rm src}}(s^\prime|s,a) - P_{\mathcal{M}_{\rm tar}}(s^\prime|s,a)| \times | Q_{\mathcal{M}_{\rm tar}}^\pi(s^\prime,a^\prime) | ds^\prime \right] \\
        &\le \dfrac{r_{\rm max}}{1-\gamma} \left[\int_\mathcal{S} |P_{\mathcal{M}_{\rm src}}(s^\prime|s,a) - P_{\mathcal{M}_{\rm tar}}(s^\prime|s,a)| ds^\prime\right] = \dfrac{2 r_{\rm max}}{1-\gamma} \left[ D_{\rm TV}(P_{\mathcal{M}_{\rm src}}(\cdot|s,a)\| P_{\mathcal{M}_{\rm tar}}(\cdot|s,a)) \right]
    \end{align*}
    Then, we get the bound for term $(a)$:
    \begin{align*}
        J_{\mathcal{M}_{\rm tar}}(\pi) - J_{\mathcal{M}_{\rm src}}(\pi_D) &\ge -\dfrac{2r_{\rm max}}{(1-\gamma)^2} \mathbb{E}_{\rho_{\mathcal{M}_{\rm src}}^{\pi_D}(s,a), s^\prime\sim P_{\mathcal{M}_{\rm src}}} \left[ D_{\rm TV}(\pi_D(\cdot|s^\prime)\| \pi(\cdot|s^\prime) \right] \\
        &- \dfrac{2 r_{\rm max}}{(1-\gamma)^2} \mathbb{E}_{\rho_{\mathcal{M}_{\rm src}}^{\pi_D}(s,a)} \left[ D_{\rm TV}(P_{\mathcal{M}_{\rm src}}(\cdot|s,a)\| P_{\mathcal{M}_{\rm tar}}(\cdot|s,a)) \right].
    \end{align*}
    Combining the bounds for term $(a)$ and term $(b)$, and we have
    \begin{align*}
        J_{\mathcal{M}_{\rm tar}}(\pi) - J_{\mathcal{M}_{\rm src}}(\pi) &\ge -\dfrac{4r_{\rm max}}{(1-\gamma)^2} \mathbb{E}_{\rho_{\mathcal{M}_{\rm src}}^{\pi_D}(s,a), s^\prime\sim P_{\mathcal{M}_{\rm src}}} \left[ D_{\rm TV}(\pi_D(\cdot|s^\prime)\| \pi(\cdot|s^\prime) \right] \\
        &- \dfrac{2 r_{\rm max}}{(1-\gamma)^2} \mathbb{E}_{\rho_{\mathcal{M}_{\rm src}}^{\pi_D}(s,a)} \left[ D_{\rm TV}(P_{\mathcal{M}_{\rm src}}(\cdot|s,a)\| P_{\mathcal{M}_{\rm tar}}(\cdot|s,a)) \right].
    \end{align*}
    Following the same procedure in the proof of Theorem \ref{theo:online} in Appendix \ref{sec:prooftheo3}, we convert the dynamics discrepancy term into the representation mismatch term, incurring the following bounds:
    \begin{align*}
        J_{\mathcal{M}_{\rm tar}}(\pi) - J_{\mathcal{M}_{\rm src}}(\pi) &\ge -\dfrac{4r_{\rm max}}{(1-\gamma)^2} \mathbb{E}_{\rho_{\mathcal{M}_{\rm src}}^{\pi_D}(s,a), s^\prime\sim P_{\mathcal{M}_{\rm src}}} \left[ D_{\rm TV}(\pi_D(\cdot|s^\prime)\| \pi(\cdot|s^\prime) \right] \\
        &- \dfrac{\sqrt{2} r_{\rm max}}{(1-\gamma)^2} \mathbb{E}_{\rho_{\mathcal{M}_{\rm src}}^{\pi_D}(s,a)} \left[ \sqrt{D_{\rm KL}(P_{\mathcal{M}_{\rm src}}(\cdot|s,a)\| P_{\mathcal{M}_{\rm tar}}(\cdot|s,a))} \right] \\
        &- \dfrac{\sqrt{2} \gamma r_{\rm max}}{(1-\gamma)^2} \mathbb{E}_{\rho_{\mathcal{M}_{\rm src}}^{\pi_D}(s,a)} \left[\sqrt{|\mathbb{H}(s_{\rm src}^\prime) - \mathbb{H}(s_{\rm tar}^\prime)|}\right].
    \end{align*}
\end{proof}

\section{Useful Lemmas}
\label{sec:usefullemma}

\begin{lemma}[Telescoping lemma]
    \label{lemma:telescopr}
    Denote $\mathcal{M}_1 = ( \mathcal{S},\mathcal{A}, P_1, r,\gamma )$ and $\mathcal{M}_2 = (\mathcal{S},\mathcal{A},P_2,r,\gamma)$ as two MDPs that only differ in their transition dynamics. Then for any policy $\pi$, we have
    \begin{equation}
        J_{\mathcal{M}_1}(\pi) - J_{\mathcal{M}_2}(\pi) = \dfrac{\gamma}{1-\gamma} \mathbb{E}_{\rho_{\mathcal{M}_1}^\pi(s,a)}\left[ \mathbb{E}_{s^\prime\sim P_1}[V_{\mathcal{M}_2}^\pi(s^\prime)] - \mathbb{E}_{s^\prime\sim P_2}[V_{\mathcal{M}_2}^\pi(s^\prime)] \right].
    \end{equation}
\end{lemma}

\begin{proof}
    This is Lemma 4.3 in \cite{luo2018algorithmic}, please check the proof there.
\end{proof}

\begin{lemma}[Extended telescoping lemma]
    \label{lemma:extendedtelescopr}
    Denote $\mathcal{M}_1 = ( \mathcal{S},\mathcal{A}, P_1, r,\gamma)$ and $\mathcal{M}_2 = (\mathcal{S},\mathcal{A},P_2,r,\gamma)$ as two MDPs that only differ in their transition dynamics. Suppose we have two policies $\pi_1, \pi_2$, we can reach the following conclusion:
    \begin{equation}
        J_{\mathcal{M}_1}(\pi_1) - J_{\mathcal{M}_2}(\pi_2) = \dfrac{1}{1-\gamma} \mathbb{E}_{\rho_{\mathcal{M}_1}^{\pi_1}(s,a)}\left[ \mathbb{E}_{s^\prime\sim P_1, a^\prime\sim\pi_1}[Q_{\mathcal{M}_2}^{\pi_2}(s^\prime,a^\prime)] - \mathbb{E}_{s^\prime\sim P_2, a^\prime\sim\pi_2}[Q_{\mathcal{M}_2}^{\pi_2}(s^\prime,a^\prime)] \right].
    \end{equation}
\end{lemma}

\begin{proof}
    This is Lemma C.2 in \cite{Xu2023CrossDomainPA}, please check the proof there.
\end{proof}

\begin{lemma}
    \label{lemma:actionlemma}
    Denote $\mathcal{M} = ( \mathcal{S},\mathcal{A}, P, r,\gamma)$ as the underlying MDP. Suppose we have two policies $\pi_1, \pi_2$, then the performance difference of these policies in the MDP gives:
    \begin{equation}
        J_{\mathcal{M}}(\pi_1) - J_{\mathcal{M}}(\pi_2) = \dfrac{1}{1-\gamma} \mathbb{E}_{\rho_{\mathcal{M}}^{\pi_1}(s,a),s^\prime\sim P}\left[ \mathbb{E}_{a^\prime\sim\pi_1}[Q_{\mathcal{M}}^{\pi_2}(s^\prime,a^\prime)] - \mathbb{E}_{a^\prime\sim\pi_2}[Q_{\mathcal{M}}^{\pi_2}(s^\prime,a^\prime)] \right].
    \end{equation}
\end{lemma}

\begin{proof}
    Similar to \cite{luo2018algorithmic}, we use a telescoping sum to prove the result. Denote $W_j$ as the expected return when deploying $\pi_1$ in the MDP $\mathcal{M}$ for the first $j$ steps and then switching to policy $\pi_2$, \emph{i.e.}, 
    \begin{equation}
        \label{eq:wj}
        W_j = \mathbb{E}_{\substack{t < j:s_t\sim P, a_t\sim\pi_1\\ t \ge j: s_t\sim P, a_t \sim \pi_2}}\left[\sum_{t=0}^\infty \gamma^t r(s_t,a_t)\right].
    \end{equation}
    By definition, it is easy to find that $W_0 = J_\mathcal{M}(\pi_2)$ and $W_\infty = J_\mathcal{M}(\pi_2)$. Next, we express the value difference as the following form:
    \begin{equation}
    \label{eq:telescopesum}
        J_{\mathcal{M}}(\pi_1) - J_{\mathcal{M}}(\pi_2) = \sum_{t=0}^\infty (W_{k+1} - W_j).
    \end{equation}
    The above term can be simplified as:
    \begin{equation}
        W_{j+1} - W_j = \gamma^{j} \mathbb{E}_{s_{j-1}\sim P,a_{j-1}\sim\pi_1}\left[ \mathbb{E}_{s_{j}\sim P,a_{j}\sim \pi_1} [Q_\mathcal{M}^{\pi_2}(s_{j},a_{j})] - \mathbb{E}_{s_{j}\sim P,a_{j}\sim \pi_2} [Q_\mathcal{M}^{\pi_2}(s_{j},a_{j})]  \right].
    \end{equation}
    Plug it back into Equation \ref{eq:telescopesum}, and we have
    \begin{align*}
        J_{\mathcal{M}}(\pi_1) - J_{\mathcal{M}}(\pi_2) &= \sum_{t=0}^\infty (W_{k+1} - W_j) \\
        &= \sum_{j=0}^\infty \gamma^j \mathbb{E}_{\rho_{\mathcal{M}}^{\pi_1}, s^\prime\sim P} \left[ \mathbb{E}_{a^\prime\sim\pi_1} [Q_\mathcal{M}^{\pi_2}(s^\prime,a^\prime)] - \mathbb{E}_{a^\prime\sim\pi_2} [Q_\mathcal{M}^{\pi_1}(s^\prime,a^\prime)] \right] \\
        &= \dfrac{1}{1-\gamma} \mathbb{E}_{\rho_{\mathcal{M}}^{\pi_1}, s^\prime\sim P} \left[ \mathbb{E}_{a^\prime\sim\pi_1} [Q_\mathcal{M}^{\pi_2}(s^\prime,a^\prime)] - \mathbb{E}_{a^\prime\sim\pi_2} [Q_\mathcal{M}^{\pi_1}(s^\prime,a^\prime)] \right].
    \end{align*}
\end{proof}

\section{Pseudocodes}
\label{sec:pseudocodes}

In this section, we provide detailed pseudocodes for online PAR and offline PAR, as shown in Algorithm \ref{alg:onlinepar} and Algorithm \ref{alg:offlinepar}.

\begin{algorithm}[tb]
\caption{Policy Adaptation by Representation Mismatch (online version)}
\label{alg:onlinepar}
{\bf Input:} Source domain $\mathcal{M}_{\rm src}$, target domain $\mathcal{M}_{\rm tar}$, target domain interaction interval $F$, batch size $N$, source domain interaction maximum step $T_{\rm max}$, reward penalty coefficient $\beta$, temperature (for SAC) $\alpha$, target update rate $\tau$.

\begin{algorithmic}[1]
\STATE Initialize policy $\pi_\phi$, value functions $\{Q_{\theta_i}\}_{i=1,2}$ and target networks $\{Q_{\theta_i^\prime}\}_{i=1,2}$, source domain replay buffer $D_{\rm src}\leftarrow \emptyset$, target domain replay buffer $D_{\rm tar}\leftarrow \emptyset$. Initialize the state encoder $f$ and state-action encoder $g$ with parameters $\psi, \xi$, respectively
\FOR{$i$ = 1, 2, ..., $T_{\rm max}$}
\STATE Collect transition $(s_{\rm src},a_{\rm src},r_{\rm src},s_{\rm src}^\prime)$ with policy $\pi_\phi$ in $\mathcal{M}_{\rm src}$
\STATE Store the transition in $D_{\rm src}, D_{\rm src}\leftarrow D_{\rm src}\cup \{(s_{\rm src},a_{\rm src},r_{\rm src},s_{\rm src}^\prime)\}$
\IF{$i$\% $F$ == 0}
\STATE Given $s_{\rm tar}$ in $\mathcal{M}_{\rm tar}$, execute $a_{\rm tar}$ using the policy $\pi_\phi$ and get $(s_{\rm tar},a_{\rm tar},r_{\rm tar},s_{\rm tar}^\prime)$
\STATE Store the transition in the replay buffer, $D_{\rm tar}\leftarrow D_{\rm tar}\cup \{(s_{\rm tar},a_{\rm tar},r_{\rm tar},s_{\rm tar}^\prime)\}$
\ENDIF
\STATE Sample $N$ transitions $d_{\rm tar} = \{(s_{j}, a_j, r_j, s_j^\prime)\}_{j=1}^N$ from $D_{\rm tar}$
\STATE Train encoders $f,g$ in the target domain by minimizing: $\frac{1}{N}\sum_{d_{\rm tar}}\left[(g_{\xi}(f_\psi(s),a) - \texttt{SG}(f_\psi(s^\prime)) )^2 \right]$
\STATE Sample $N$ transitions $d_{\rm src} = \{(s_{j}, a_j, r_j, s_j^\prime)\}_{j=1}^N$ from $D_{\rm src}$
\STATE Modify source domain rewards into $\hat{r}_{\rm src} = r_{\rm src} - \beta \times \left[g_\xi(f_\psi(s_{\rm src}),a_{\rm src}) - f_\psi(s_{\rm src}^\prime) \right]^2$
\STATE Calculate target values $y = r + \gamma\left[\min_{i=1,2}Q_{\theta_i^\prime}(s^\prime,a^\prime) - \alpha \log\pi_\phi(a^\prime|s^\prime)\right], a^\prime\sim\pi_\phi(\cdot|s^\prime)$
\STATE Update critics by minimizing $\frac{1}{2N}\sum_{d_{\rm src}\cup d_{\rm tar}}(Q_{\theta_i} - y)^2, i\in\{1,2\}$
\STATE Update actor by maximizing $\frac{1}{2N}\sum_{d_{\rm src}\cup d_{\rm tar}, a\sim\pi_\phi(\cdot|s)} \left[ \min_{i=1,2}Q_{\theta_i}(s,a) - \alpha\log\pi_\phi(\cdot|s) \right]$
\STATE Update target networks: $\theta_i^\prime\leftarrow \tau\theta_i + (1-\tau)\theta_i^\prime$
\ENDFOR
\end{algorithmic}
\end{algorithm}

\begin{algorithm}[tb]
\caption{Policy Adaptation by Representation Mismatch (offline version)}
\label{alg:offlinepar}
{\bf Input:} Source domain $\mathcal{M}_{\rm src}$, target domain $\mathcal{M}_{\rm tar}$, target domain interaction interval $F$, batch size $N$, maximum gradient step $T_{\rm max}$, reward penalty coefficient $\beta$, normalization coefficient $\nu$, temperature (for SAC) $\alpha$, target update rate $\tau$. Source domain offline dataset $\mathcal{D}_{\rm off}$

\begin{algorithmic}[1]
\STATE Initialize policy $\pi_\phi$, value functions $\{Q_{\theta_i}\}_{i=1,2}$ and target networks $\{Q_{\theta_i^\prime}\}_{i=1,2}$, source domain replay buffer $D_{\rm src}\leftarrow D_{\rm off}$, target domain replay buffer $D_{\rm tar}\leftarrow \emptyset$. Initialize the state encoder $f$ and state-action encoder $g$ with parameters $\psi, \xi$, respectively
\FOR{$i$ = 1, 2, ..., $T_{\rm max}$}
\IF{$i$\% $F$ == 0}
\STATE Given $s_{\rm tar}$ in $\mathcal{M}_{\rm tar}$, execute $a_{\rm tar}$ using the policy $\pi_\phi$ and get $(s_{\rm tar},a_{\rm tar},r_{\rm tar},s_{\rm tar}^\prime)$
\STATE Store the transition in the replay buffer, $D_{\rm tar}\leftarrow D_{\rm tar}\cup \{(s_{\rm tar},a_{\rm tar},r_{\rm tar},s_{\rm tar}^\prime)\}$
\ENDIF
\STATE Sample $N$ transitions $d_{\rm tar} = \{(s_{j}, a_j, r_j, s_j^\prime)\}_{j=1}^N$ from $D_{\rm tar}$
\STATE Train encoders $f,g$ in the target domain by minimizing: $\frac{1}{N}\sum_{d_{\rm tar}}\left[(g_{\xi}(f_\psi(s),a) - \texttt{SG}(f_\psi(s^\prime)) )^2 \right]$
\STATE Sample $N$ transitions $d_{\rm src} = \{(s_{j}, a_j, r_j, s_j^\prime)\}_{j=1}^N$ from $D_{\rm src}$
\STATE Modify source domain rewards into $\hat{r}_{\rm src} = r_{\rm src} - \beta \times \left[g_\xi(f_\psi(s_{\rm src}),a_{\rm src}) - f_\psi(s_{\rm src}^\prime) \right]^2$
\STATE Calculate target values $y = r + \gamma\left[\min_{i=1,2}Q_{\theta_i^\prime}(s^\prime,a^\prime) - \alpha \log\pi_\phi(a^\prime|s^\prime)\right], a^\prime\sim\pi_\phi(\cdot|s^\prime)$
\STATE Update critics by minimizing $\frac{1}{2N}\sum_{d_{\rm src}\cup d_{\rm tar}}(Q_{\theta_i} - y)^2, i\in\{1,2\}$
\STATE Update actor by maximizing $\frac{\lambda}{2N}\sum_{d_{\rm src}\cup d_{\rm tar}, a\sim\pi_\phi(\cdot|s)} \left[ \min_{i=1,2}Q_{\theta_i}(s,a) - \alpha\log\pi_\phi(\cdot|s) \right] - \frac{1}{N}\sum_{d_{\rm src}}(a - \tilde{a})^2$, where $\tilde{a}\sim\pi_\phi(\cdot|s), \lambda = \nicefrac{\nu}{\frac{1}{2N}\sum_{d_{\rm src}\cup d_{\rm tar}} \min_{i=1,2} Q_{\theta_i}(s,a)}$
\STATE Update target networks: $\theta_i^\prime\leftarrow \tau\theta_i + (1-\tau)\theta_i^\prime$
\ENDFOR
\end{algorithmic}
\end{algorithm}

\section{Experimental Details and Hyperparameter Setup}

In this section, we describe the detailed experimental setup as well as the hyperparameter setup used in this work. To ensure reproducibility, we include the codes of PAR in the supplementary material, and will open source our codes upon acceptance. 

\subsection{Environment Setting}
\label{sec:environmentsetting}

We adopt the same environments proposed in \cite{Xu2023CrossDomainPA} without any modification. We employ four widely used environments from OpenAI Gym \cite{Brockman2016OpenAIG}, \emph{HalfCheetah-v2}, \emph{Hopper-v2}, \emph{Walker2d-v2}, \emph{Ant-v3}, as source domains. To simulate dynamics discrepancies, we consider kinematic shifts and morphology shifts between the source domain and the target domain. This results in a total of 8 target domains. Kinematic shifts indicate that some joints of the simulated robot are \emph{broken}, while morphology mismatch means that there are some morphological differences between the simulated robots in the two domains. We explicate the detailed modifications below.

\textbf{\emph{halfcheetah (broken back thigh)}}: The rotation angle of the joint on the thigh of the Cheetah robot's back leg is modified from $[-0.52, 1.05]$ to $[-0.0052, 0.0105]$.

\textbf{\emph{hopper (broken joints)}}: The rotation angles of the head joint and the foot joint are modified from $[-150,0], [-45, 45]$ to $[-0.15, 0], [-18, 18]$, respectively.

\textbf{\emph{walker (broken right foot)}}: The rotation angle of the foot joint on the robot's right leg is modified from $[-45, 45]$ to $[-0.45, 0.45]$.

\textbf{\emph{ant (broken hips)}}: The rotation angle of the joints on the hip of two legs are modified from $[-30, 30]$ to $[-0.3,0.3]$

\textbf{\emph{halfcheetah (no thighs)}}: The sizes of the back thigh and the forward thigh are reduced as shown below:

\begin{lstlisting}[language=python]
# back thigh
<geom fromto="0 0 0 -0.0001 0 -0.0001" name="bthigh" size="0.046" type="capsule"/>
<body name="bshin" pos="-0.0001 0 -0.0001">
# forward thigh
<geom fromto="0 0 0 0.0001 0 0.0001" name="fthigh" size="0.046" type="capsule"/>  
<body name="fshin" pos="0.0001 0 0.0001">
\end{lstlisting}

\textbf{\emph{hopper (big head)}}: The head size of the robot is modified as shown below:

\begin{lstlisting}[language=python]
# head size
<geom friction="0.9" fromto="0 0 1.45 0 0 1.05" name="torso_geom" size="0.125" type="capsule"/>
\end{lstlisting}

\textbf{\emph{walker (no right foot)}}: The thigh on the right leg of the robot is modified as the following:

\begin{lstlisting}[language=python]
# right leg
<body name="thigh" pos="0 0 1.05">
  <joint axis="0 -1 0" name="thigh_joint" pos="0 0 1.05" range="-150 0" type="hinge"/>
  <geom friction="0.9" fromto="0 0 1.05 0 0 1.045" name="thigh_geom" size="0.05" type="capsule"/>
  <body name="leg" pos="0 0 0.35">
    <joint axis="0 -1 0" name="leg_joint" pos="0 0 1.045" range="-150 0" type="hinge"/>
    <geom friction="0.9" fromto="0 0 1.045 0 0 0.3" name="leg_geom" size="0.04" type="capsule"/>
    <body name="foot" pos="0.2 0 0">
      <joint axis="0 -1 0" name="foot_joint" pos="0 0 0.3" range="-45 45" type="hinge"/>
      <geom friction="0.9" fromto="-0.0 0 0.3 0.2 0 0.3" name="foot_geom" size="0.06" type="capsule"/>
    </body>
  </body>
</body>
\end{lstlisting}

\textbf{\emph{ant (short feet)}}: The size of the ant robot's feet on its front two legs are modified into the following parameters:

\begin{lstlisting}[language=python]
# leg 1
<geom fromto="0.0 0.0 0.0 0.1 0.1 0.0" name="left_ankle_geom" size="0.08" type="capsule"/>
# leg 2
<geom fromto="0.0 0.0 0.0 -0.1 0.1 0.0" name="right_ankle_geom" size="0.08" type="capsule"/>
\end{lstlisting}

For more details, one could check the \texttt{xml} files in the supplementary material.

\begin{figure}
    \centering
    \includegraphics[width=0.95\linewidth]{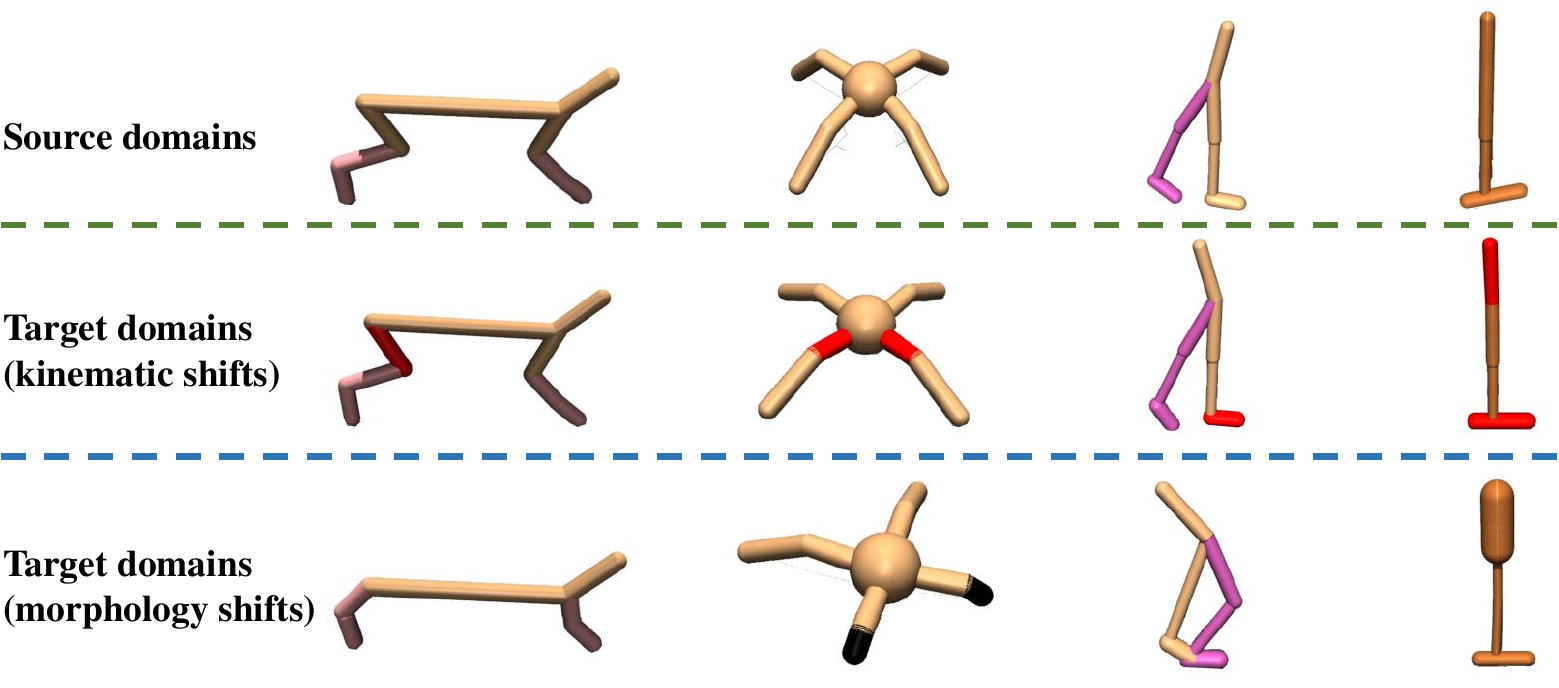}
    \caption{\textbf{Graphical illustration of the evaluated environments.} The source domains (\emph{top}) are well-functional simulated robots from Gym, the target domains either have kinematic shifts (\emph{middle}) or morphology shifts (\emph{bottom}) compared to the source domains.}
    \label{fig:environments}
\end{figure}

\subsection{Implementation Details}
\label{sec:implementationdetails}

In this subsection, we provide implementation details as well as an introduction of the baselines adopted in this work and the PAR algorithm. When the source domain is online, we consider the following baselines for comparison: SAC-tar \cite{haarnoja2018softactorcritic}, DARC \cite{eysenbach2021offdynamics}, DARC-weight, VGDF \cite{Xu2023CrossDomainPA}, and SAC-tune. We list the detailed implementation details of these methods below.

\textbf{SAC-tar}: We train an SAC agent solely in the target domain for $10^5$ environmental steps, with its hyperparameter setup specified in Table \ref{tab:hyperparametersetup}. We do not use the temperature auto-tuned SAC, but keep the temperature coefficient $\alpha=0.2$ fixed. This applies to all of the other baselines.

\textbf{SAC-tune}: We train an SAC agent first in the source domain for $10^6$ environmental steps, and then fine-tune its policy by interacting with the target domain for another $10^5$ timesteps. We use the same hyperparameters as SAC-tar.

\textbf{DARC}: We follow the original paper to train two domain classifiers $q_{{\theta}_{\rm SAS}}({\rm{target}}|s_t,a_t,s_{t+1}), q_{\theta_{\rm SA}}({\rm{target}}|s_t,a_t)$ parameterized by $\theta_{\rm SAS}$ and $\theta_{\rm SA}$, respectively. These domain classifiers are optimized via the cross-entropy loss:
\begin{align*}
    \mathcal{L}(\theta_{\rm SAS}) = \mathbb{E}_{\mathcal{D}_{\rm tar}}\left[ \log q_{\theta_{\rm SAS}}({\rm target}| s_t,a_t,s_{t+1}) \right] + \mathbb{E}_{\mathcal{D}_{\rm src}}\left[ \log (1-q_{\theta_{\rm SAS}}({\rm target}|s_t,a_t,s_{t+1})) \right], \\
    \mathcal{L}(\theta_{\rm SA}) = \mathbb{E}_{\mathcal{D}_{\rm tar}}\left[ \log q_{\theta_{\rm SA}}({\rm target}| s_t,a_t) \right] + \mathbb{E}_{\mathcal{D}_{\rm src}}\left[ \log (1-q_{\theta_{\rm SA}}({\rm target}|s_t,a_t)) \right],
\end{align*}
where $D_{\rm src}, D_{\rm tar}$ are replay buffers of the source domain and the target domain, respectively. Following the original paper, we use Gaussian standard deviation $\sigma = 1$ for training the domain classifiers. We experimentally find that DARC is quite sensitive to the standard deviation $\sigma$, \emph{e.g.}, if one sets $\sigma=0.1$, DARC exhibits very poor performance on almost all tasks. We do not modify this value and keep it fixed across all runs. Then, DARC compensates the source domain rewards by estimating the dynamics gap with the form: $\log \frac{P_{\mathcal{M}_{\rm tar}}(s_{t+1}|s_t,a_t)}{P_{\mathcal{M}_{\rm src}}(s_{t+1}|s_t,a_t)}$. By approximating this term with the trained encoders, DARC estimates the reward correction term $\delta r$ by
\begin{equation}
\label{eq:darcrewardformula}
    \delta r(s_t,a_t) = - \log\dfrac{q_{\theta_{\rm SAS}}({\rm target}|s_t,a_t,s_{t+1}) q_{\theta_{\rm SA}}({\rm source}|s_t,a_t)}{q_{\theta_{\rm SAS}}({\rm source}|s_t,a_t,s_{t+1}) q_{\theta_{\rm SA}}({\rm target}|s_t,a_t)}.
\end{equation}
The source domain rewards are formally modified via:
\begin{equation}
    \label{eq:darcrewardmodify}
    \hat{r}_{\rm src}^{\rm DARC} = r_{\rm src}(s_t,a_t) - \beta\times\delta_t,
\end{equation}
where $\beta\in\mathbb{R}$ is the reward penalty coefficient. Note that Equation \ref{eq:darcrewardmodify} is slightly different from the original paper in the following aspects: (1) DARC paper adopts $\beta=1$ by default and does not tune this hyperparameter, while we search the best $\beta$ across $\{0.1,0.5,1.0,2.0\}$ for online experiments; (2) we adopt a negative reward correction term, \emph{i.e.}, $r - \delta r$, whereas DARC paper has the form $r + \delta r$. We use this form to ensure consistency between PAR and DARC in reward correction, and for the benefit of reward penalty comparison illustrated in Figure \ref{fig:darcparrewardcompare} and Figure \ref{fig:rewardcompappendix}. We provide DARC's the best hyperparameter $\beta$ for each task in Table \ref{tab:parametereveryenvironment}. We adopt the default hyperparameter setup from the authors (\href{https://github.com/google-research/google-research/tree/master/darc}{https://github.com/google-research/google-research/tree/master/darc}). Moreover, we follow the instruction in Appendix E of the DARC paper and warmup the algorithm without $\delta r$ for the first $10^5$ steps.

\textbf{DARC-weight}: This variant generally resembles vanilla DARC, except that it does not perform reward correction for the source domain data, but utilizes that estimated dynamics gap as an importance sampling term for critic updates, \emph{i.e.},
\begin{equation}
    \label{eq:darcweight}
    \mathcal{L}_{\rm critic} = \mathbb{E}_{(s,a,r,s^\prime)\sim\mathcal{D}_{\rm src}}\left[ \omega(s,a,s^\prime)\left(Q_{\theta_i}(s,a) - y\right)^2 \right], i\in\{1,2\},
\end{equation}
where
\begin{equation}
\label{eq:omega}
    \omega(s,a,s^\prime) = \frac{q_{\theta_{\rm SAS}}({\rm target}|s,a,s^\prime) q_{\theta_{\rm SA}}({\rm source}|s,a)}{q_{\theta_{\rm SAS}}({\rm source}|s,a,s^\prime) q_{\theta_{\rm SA}}({\rm target}|s,a)}.
\end{equation}
To ensure that the importance sampling weight $\omega(s,a,s^\prime)$ lies in a valid range, we clip its value to the range of $[1e^{-4}, 1]$ for training stability.

\textbf{VGDF}: VGDF is constructed based on the theoretical results that the performance bound of a policy in the source domain and the target domain is controlled by value difference, \emph{i.e.}, 
\begin{equation}
    J_{\mathcal{M}_{\rm tar}}(\pi) \ge J_{\mathcal{M}_{\rm src}}(\pi) - \dfrac{\gamma}{1-\gamma}\mathbb{E}_{\rho_{\mathcal{M}_{\rm src}}^\pi}\left[ \left| \mathbb{E}_{P_{\mathcal{M}_{\rm src}}}[V_{\mathcal{M}_{\rm tar}}^\pi(s^\prime)] - \mathbb{E}_{P_{\mathcal{M}_{\rm tar}}}[V_{\mathcal{M}_{\rm tar}}^\pi(s^\prime)] \right| \right].
\end{equation}
Then, VGDF decides to filter samples in the source domain that share similar value estimates as those in the target domain. To that end, it trains an ensemble of dynamics model akin to model-based RL \cite{Janner2019WhenTT, Pan2020TrustTM, Qiao2023ThePB, Buckman2018SampleEfficientRL} in the \emph{original state-action space} of the target domain to predict the next state that follows the transition dynamics of the target domain given source domain data $(s_{\rm src}, a_{\rm src})$. Then it measures the mean and variance of the value ensemble $\{Q(s_i^\prime,a_i^\prime)\}_{i=1}^M$ to construct a Gaussian distribution, where $s_i^\prime$ is the predicted next state, $a_i^\prime$ is sampled from the policy, and $M$ is the ensemble size. After that, the authors utilize rejection sampling to select a fixed percentage of source domain data with the highest likelihood estimation and share them with the target domain. We use the official implementation of VGDF (\href{https://github.com/Kavka1/VGDF}{https://github.com/Kavka1/VGDF}) without any modification. We set the data selection ratio in VGDF as 25\%. VGDF also adopts SAC as the base algorithm, with its temperature set to be $0.2$. VGDF trains an extra exploration policy for better exploration (while PAR does not). Following the original implementation, we warm-start VGDF by disabling rejection sampling (\emph{i.e.}, accept all transitions from the source domain) for the first $10^5$ timesteps. We generally can reproduce the reported performance of VGDF.

\textbf{PAR}: Different from the above methods, PAR detects the dynamics mismatch by capturing the representation mismatch, \emph{i.e.}, the representation deviation between the source domain state-action pair and its subsequent next state using the state encoder $f$ and state-action encoder $g$ trained only in the target domain. Note that different from VGDF, we only train \emph{one single} state encoder along with \emph{one single} state-action encoder, which we find can already incur satisfying performance and suitable reward penalty. The encoders are updated via Equation \ref{eq:representationobjective}. We compensate the source domain rewards by measuring the representation deviation, as shown in Equation \ref{eq:rewardmodification}. The detailed hyperparameter setup for PAR is available in Table \ref{tab:hyperparametersetup}. We use the same batch size of source domain data and target domain data for training. Since the representation deviation is strongly correlated with the environment itself, \emph{e.g.}, the state space, the action space, and the reward function, we believe it is understandable that the best penalty coefficient $\beta$ differs among different evaluated online tasks. We sweep across $\beta\in\{0.1,0.5,1.0,2.0\}$ and report the adopted $\beta$ for each task in Table \ref{tab:parametereveryenvironment}. 

When the source domain is offline, we consider baseline methods of CQL-0 \cite{Kumar2020ConservativeQF}, CQL+SAC, H2O \cite{niu2022when}, and VGDF+BC \cite{Xu2023CrossDomainPA}. The corresponding implementation details can be found below.

\textbf{CQL-0}: CQL is a well-known offline RL algorithm. CQL-0 denotes that we train a CQL agent merely on the offline source domain dataset and transfer the learned policy to the target domain in a zero-shot manner. We use the public implementation of CQL (\href{https://github.com/tinkoff-ai/CORL}{https://github.com/tinkoff-ai/CORL}) and train it for $10^6$ gradient steps.

\textbf{CQL+SAC}: This baseline leverages both offline source domain data and online target domain transitions for learning a policy. Since learning from offline data requires conservatism, while learning from online samples does not, we train critics by updating source domain data with the CQL loss while the online target domain data with the SAC loss, \emph{i.e.},
\begin{equation}
    \mathcal{L}_{\rm critic} = \mathbb{E}_{D_{\rm src}\cup D_{\rm tar}}\left[ (Q_{\theta_i}(s,a) - y)^2 \right] + \beta_{\rm CQL} \left( \mathbb{E}_{s\sim D_{\rm src},\tilde{a}\sim\pi_\phi(\cdot|s)}[Q_{\theta_i}(s,\tilde{a})] - \mathbb{E}_{D_{\rm src}}[Q_{\theta_i}(s,a)] \right), i\in\{1,2\},
\end{equation}
where $\beta_{\rm CQL}$ is the hyperparameter, and we use $\beta_{\rm CQL} = 10.0$ (which is the same as CQL-0). Note that we sample the same batch size ($128$) of both source domain data and target domain for update at each gradient step. We train CQL+SAC for $10^6$ gradient steps, with $10^5$ interactions with the target domain.

\textbf{H2O}: H2O can be viewed as an offline version of \emph{DARC-weight} algorithm, which also trains domain classifiers to estimate dynamics gap, which further serves as importance sampling weights for critic optimization. It additionally combines CQL loss to inject conservatism. To be specific, its critic objective function can be written as:
\begin{equation}
\begin{split}
    \mathcal{L}_{\rm critic} &= \mathbb{E}_{D_{\rm tar}}\left[ (Q_{\theta_i}(s,a) - y)^2 \right] + \mathbb{E}_{D_{\rm src}}\left[ \omega(s,a,s^\prime)(Q_{\theta_i}(s,a) - y)^2 \right] \\
    &\qquad\qquad + \beta_{\rm CQL} \left( \mathbb{E}_{s\sim D_{\rm src},\tilde{a}\sim\pi_\phi(\cdot|s)}[Q_{\theta_i}(s,\tilde{a})] - \mathbb{E}_{D_{\rm src}}[Q_{\theta_i}(s,a)] \right), i\in\{1,2\},
\end{split}
\end{equation}
where $\omega(s,a,s^\prime)$ is evaluated as in Equation \ref{eq:omega}. We use the default configurations of hyperparameters specified in the official codebase (\href{https://github.com/t6-thu/H2O}{https://github.com/t6-thu/H2O}), except that we use $\beta_{\rm CQL}=10.0$ since we experimentally find that the recommended $\beta_{\rm CQL}=0.01$ from the authors incurs very poor performance on all of the datasets. We train H2O for $10^6$ gradient steps, and allow the policy to gather data in the target domain every 10 steps. 

\textbf{VGDF+BC}: VGDF+BC generally has the same hyperparameter setup as VGDF, except that we incorporate a behavior cloning term into its policy training, akin to offline PAR. Its actor generally follows the same way of updating as Equation \ref{eq:offlineactorloss}. We take the results of VGDF+BC on six medium-level datasets from its paper directly, where one can see that it actually uses different hyperparameter setups on different tasks, \emph{i.e.}, $\nu$. For other datasets, we set $\nu=5$ by following the instructions from the original paper. We train VGDF+BC for $10^6$ gradient steps, with $10^5$ interactions with the target domain.

\textbf{PAR}: Offline PAR differs from its online variant in that an additional behavior cloning term is involved. We generally adopt fixed reward penalty coefficient $\beta$ and normalization parameter $\nu$ across all tasks, as depicted in Table \ref{tab:parametereveryenvironment}. We train PAR for $10^6$ gradient steps, and let it interact with the target domain every 10 gradient steps.

\subsection{Hyperparameter Setup}
In this part, we list the detailed hyperparameter setup for PAR and baseline methods in Table \ref{tab:hyperparametersetup}. We also provide the adopted key hyperparameters given both the online source domain and the offline source domain in Table \ref{tab:parametereveryenvironment}.

\begin{table*}
\centering
\caption{Detailed hyperparameter setup for PAR and baseline methods on the evaluated tasks.}
\label{tab:hyperparametersetup}
\begin{tabular}{lrr}
\toprule
\textbf{Hyperparameter}  & \textbf{Value} \quad \\
\midrule
Shared & \\
\qquad Actor network  & \qquad  $(256,256)$ \\
\qquad Critic network & \qquad $(256,256)$ \\
\qquad Batch size     &\qquad   $256$ for SAC-tar, CQL-0, and $128$ for others \\
\qquad Learning rate  & \qquad  $3\times 10^{-4}$ \\
\qquad Optimizer & \qquad Adam \cite{KingmaB14adam} \\
\qquad Discount factor & \qquad $0.99$ \\
\qquad Replay buffer size & \qquad $10^6$  \\
\qquad Warmup steps & \qquad $0$ for PAR and $10^5$ for others \\
\qquad Nonlinearity & \qquad ReLU \\
\qquad Target update rate & \qquad $5\times 10^{-3}$ \\
\qquad Temperature coefficient & \qquad $0.2$ \\
\qquad Maximum log std & \qquad $2$ \\
\qquad Minimum log std & \qquad $-20$ \\
\qquad Target domain interaction interval & \qquad $10$ \\
\midrule
DARC, DARC-weight, H2O & \\
\qquad Classifier Network & \qquad $(256,256)$ \\
\midrule
CQL-0, CQL+SAC, H2O & \\
\qquad CQL penalty coefficient $\beta_{\rm CQL}$ & \qquad $10.0$ \\
\midrule
VGDF, VGDF+BC & \\
\qquad Dynamics model network & \qquad $(200,200,200,200,200)$ \\
\qquad Ensemble size & \qquad $7$ \\
\qquad Data selection ratio & \qquad $25$\% \\
\qquad Normalization coefficient $\nu$ & \qquad $5.0$ \\
\midrule
VGDF & \\
\qquad Exploration policy network & \qquad $(256,256)$ \\
\midrule
PAR & \\
\qquad Encoder Network & \qquad $(256, 256)$ \\
\qquad Representation dimension & \qquad $256$ \\
\qquad Nomralization coefficient $\nu$ & \qquad $5.0$ \\
\bottomrule
\end{tabular}
\end{table*}

\begin{table*}
\centering
\caption{Adopted hyperparameters for PAR and baseline method DARC on evaluated environments.}
\label{tab:parametereveryenvironment}
\begin{tabular}{l|c|c|c|c}
\toprule
Task Name & PAR (online) $\beta$ & DARC (online) $\beta$ & PAR (offline) $\beta$ & PAR (offline) $\nu$ \\
\midrule
halfcheetah (broken back thigh) & 1.0 & 2.0 & 1.0 & 5.0 \\
halfcheetah (no thighs) & 2.0 & 0.5 & 1.0 & 5.0 \\
hopper (broken joints) & 0.5 & 2.0 & 1.0 & 5.0 \\
hopper (big head) & 0.5 & 1.0 & 1.0 & 5.0  \\
walker (broken right foot) & 0.5 & 1.0 & 1.0 & 5.0 \\
walker (no right thigh) & 0.5 & 1.0 & 1.0 & 5.0 \\
ant (broken hips) & 0.1 & 1.0 & 1.0 & 5.0 \\
ant (short feet) &  0.1 & 0.1 & 1.0 & 5.0 \\
\bottomrule
\end{tabular}
\end{table*}

\section{Extended Experimental Results}
\label{sec:extendedexperiments}

In the main text, we cannot include all of our experiments due to the space limit. In this section, we provide more experimental results concerning on parameter study (\emph{i.e.}, more experiments with the online source domain and results given the offline source domain), and the reward penalty comparison between DARC and PAR. We believe these are helpful to better understand the effect of the key hyperparameters in PAR and further validate the advantages of PAR against DARC.

\subsection{Wider Parameter Study}

\begin{figure}[!t]
    \centering
    \subfigure[Penalty coefficient $\beta$.]{
    \label{fig:penaltycoefficientappendix}
    \includegraphics[width=0.47\linewidth]{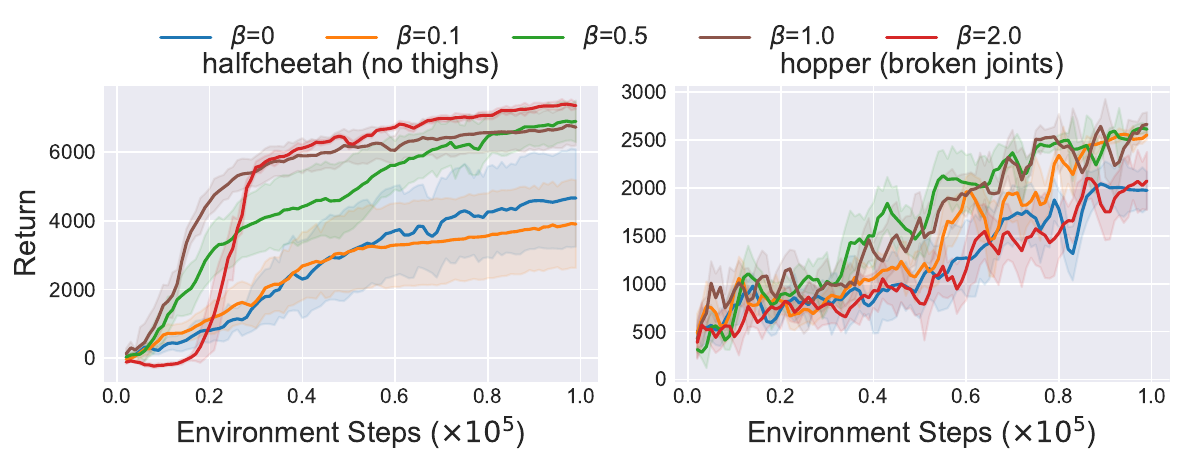}
    }\hspace{0mm}
    \subfigure[Target domain interaction interval $F$.]{
    \label{fig:targetinteractionfrequencyappendix}
    \includegraphics[width=0.47\linewidth]{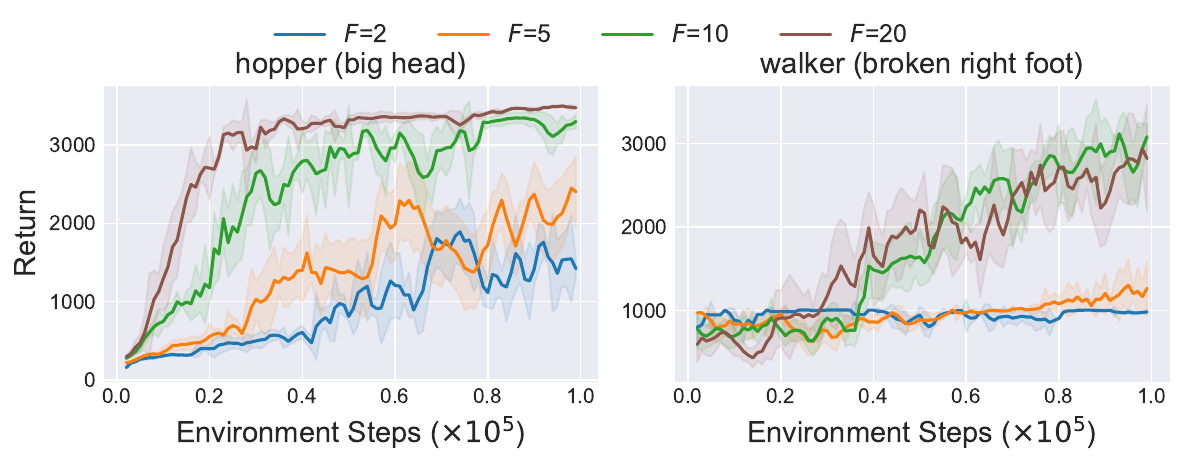}
    }
    \caption{\textbf{Extended parameter study of (a) reward penalty coefficient $\beta$, (b) target domain interaction interval $F$ on wider environments.} These curves show the performance of the policy in the target domain. The reported results are averaged across 5 different random seeds and the shaded region represents the standard deviation.}
    \label{fig:parameterstudycoefficientneighborappendix}
\end{figure}

We first include more results of the parameter study of reward penalty coefficient $\beta$ and target domain interaction interval $F$, as illustrated in Figure \ref{fig:penaltycoefficientappendix} and Figure \ref{fig:targetinteractionfrequencyappendix}, respectively. Note that the results are conducted over the online PAR algorithm. For the reward penalty coefficient $\beta$, setting $\beta=0$ makes PAR degenerate into VGDF with the data selection ratio 0\%. We find that setting $\beta=0$ (\emph{i.e.}, no reward modification for source domain data) generally does not incur a good performance, which is consistent with the results in Figure \ref{fig:penaltycoefficient} of the main text. Note that these tasks also have different optimal $\beta$.

As for the target domain interaction interval $F$, one can see that on other tasks, larger $F$ also results in a better performance. This indicates that the amount of source domain data is critical for PAR, and more data from the source domain can boost the performance of PAR in the target domain.

Next, we investigate how the hyperparameters affect the performance of PAR given an offline source domain. We are interested in the reward penalty coefficient $\beta$ and the normalization parameter $\nu$. We sweep across $\beta\in\{0, 0.1,0.5,1.0,2.0\}$ and $\nu\in\{1.0,2.5,5.0,10.0\}$. We summarize the analysis and experimental results below.

\begin{figure}[!t]
    \centering
    \subfigure[Comparison of different reward penalty coefficient $\beta$ given medium-level datasets.]{
    \label{fig:mediumbeta}
    \includegraphics[width=0.95\linewidth]{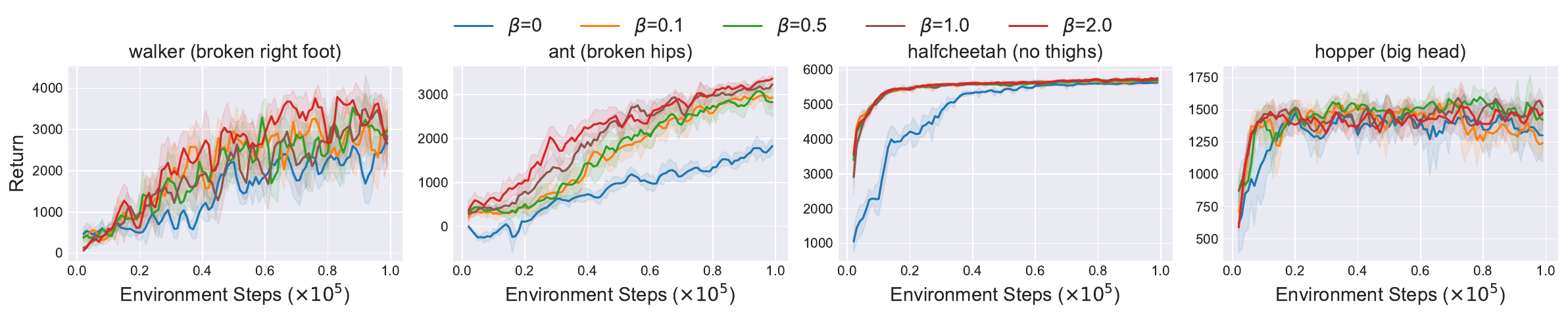}
    }\hspace{0mm}
    \subfigure[Comparison of different reward penalty coefficient $\beta$ given medium-expert-level datasets.]{
    \label{fig:mediumexpertbeta}
    \includegraphics[width=0.95\linewidth]{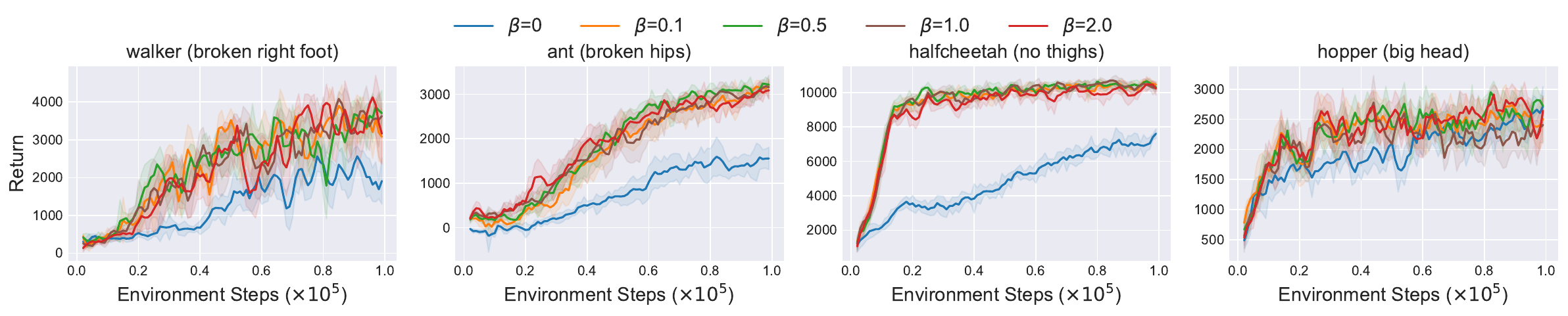}
    }
    \caption{\textbf{Extended parameter study of reward penalty coefficient $\beta$ given medium, medium-expert level source domain datasets.} We report the average return over 5 different runs along with the standard deviation of the policy in the target domain.}
    \label{fig:appendixbeta}
\end{figure}

\textbf{Reward penalty coefficient $\beta$ in offline PAR.} We consider two quality levels, \emph{medium} and \emph{medium-expert}, from D4RL datasets as the source domain, and run experiments on four tasks, two with kinematic shifts (\emph{walker (broken right foot)}, \emph{ant (broken hips)}) and two with morphology shifts (\emph{halfcheetah (no thighs)}, \emph{hopper (big head)}). We present the results in Figure \ref{fig:appendixbeta}. We find that setting $\beta=0$ incurs worse performance on most of the tasks, regardless of whether medium-level datasets or medium-expert-level datasets are provided. This further demonstrates the necessity of reward modification by PAR and highlights the effectiveness of our method. We also observe that the performance of offline PAR is generally better with higher quality datasets. Meanwhile, it still holds that the best penalty coefficient $\beta$ differs in different environments. For all of our offline experiments, we set $\beta=1$ by default and do not tune this value.

\begin{figure}[!t]
    \centering
    \subfigure[Comparison of different normalization coefficient $\nu$ given medium-level datasets.]{
    \label{fig:mediumnu}
    \includegraphics[width=0.95\linewidth]{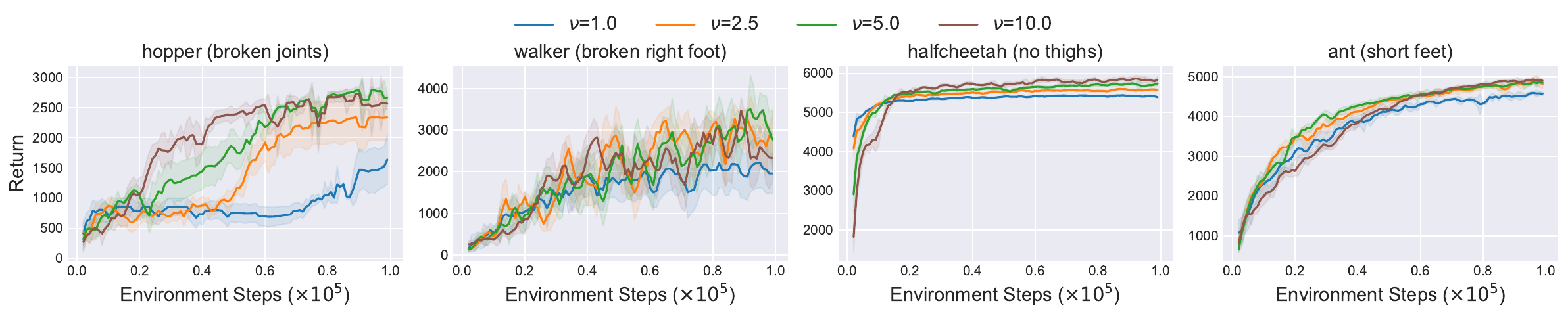}
    }\hspace{0mm}
    \subfigure[Comparison of different normalization coefficient $\nu$ given medium-expert-level datasets.]{
    \label{fig:mediumexpertnu}
    \includegraphics[width=0.95\linewidth]{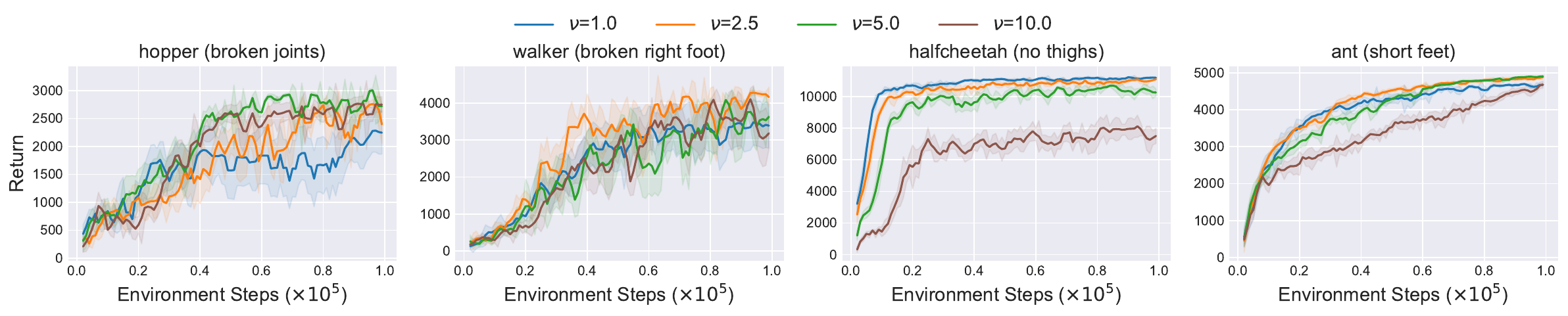}
    }
    \caption{\textbf{Extended parameter study of normalization coefficient $\nu$ under medium, medium-expert level source domain datasets.} The results depict the mean performance of the policy in the target domain with 5 different random seeds. The shaded region is the standard deviation.}
    \label{fig:appendixnu}
\end{figure}

\textbf{Normalization coefficient $\nu$ in offline PAR.} $\nu$ controls the balance between the behavior cloning term and the term that maximizes the Q-value. Intuitively, a larger $\nu$ tends to bring more conservatism, encouraging the learned policy to be close to the behavior policy in the source domain dataset. We run experiments on four tasks that are made up of two tasks with kinematic shifts (\emph{hopper (broken joints)}, \emph{walker (broken right foot)}) and two tasks with morphology shifts (\emph{halfcheetah (no thighs)}, \emph{ant (short feet)}). The results can be found in Figure \ref{fig:appendixnu}. We can see that PAR is robust to $\nu$ on tasks like \emph{walker (broken right foot)}. While it turns out that on tasks like \emph{hopper (broken joints)}, a too small $\nu$ is not preferred, and on tasks like \emph{halfcheetah (no thighs)}, a too large $\nu$ does not ensure good performance. We therefore adopt $\nu=5.0$ for all of the offline experiments to seek a trade-off.

\subsection{Wider Evidence on Reward Penalty Comparison between DARC and PAR}

\begin{figure}
    \centering
    \includegraphics[width=0.95\linewidth]{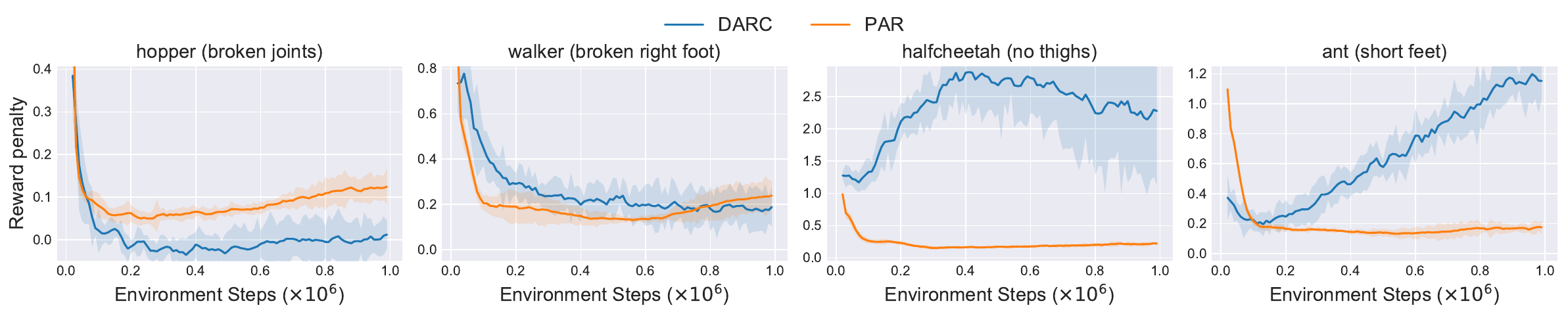}
    \caption{\textbf{Extensive comparison on reward penalty produced by DARC and PAR.} We record the mean reward penalty calculated in a sampled batch during the training of each algorithm. The environmental steps here denote the step in the source domain. The reported curves are further averaged across 5 independent runs and the shaded region captures the standard deviation.}
    \label{fig:rewardcompappendix}
\end{figure}

In this part, we provide more evidence that PAR is better than DARC. Since DARC is an online algorithm, we conduct experiments with the online source domains. We run experiments on wider environments, including two environments with kinematic mismatch (\emph{hopper (broken joints)}, \emph{walker (broken right foot)}) and two tasks with morphology mismatch (\emph{halfcheetah (no thighs)}, \emph{ant (short feet)}). We summarize the comparison results in Figure \ref{fig:rewardcompappendix}. Based on the curves, we find that on many tasks (\emph{e.g.}, \emph{halfcheetah (no thighs)}), the reward penalty given by DARC is quite large and even becomes larger with more environment steps, indicating that DARC can be overly pessimistic and the classifiers may fail to produce suitable reward penalties to compensate source domain data. On the \emph{hopper (broken joints)} task, however, DARC gives penalties that quite approach 0, and fails to inform the agent of the dynamics difference between the source domain and the target domain. Instead, we observe that on all of the evaluated tasks, the reward penalty given by PAR gradually converges to a small number (but not 0). Despite that at the initial stage, the collected samples from the source domain and the target domain may have large discrepancies, our method can fully exploit these data and successfully find dynamics-consistent behaviors and transitions later on. We believe these further verify that PAR is a better choice than DARC.

\section{Compute Infrastructure}

In Table \ref{tab:computing}, we list the compute infrastructure that we use to run all of the algorithms.

\begin{table}[htb]
\caption{Compute infrastructure.}
\label{tab:computing}
\vspace{2mm}
\centering
\begin{tabular}{c|c|c}
\toprule
\textbf{CPU}  & \textbf{GPU} & \textbf{Memory} \\
\midrule
AMD EPYC 7452  & RTX3090$\times$8 & 288GB \\
\bottomrule
\end{tabular}
\end{table}

\end{document}